\documentclass{article}

\usepackage{microtype}
\usepackage{graphicx}
\usepackage{subfigure}
\usepackage{booktabs} 

\usepackage{hyperref}

\usepackage[accepted]{icml2018}

\icmltitlerunning{World Models}

\begin{document}

\twocolumn[
\icmltitle{World Models}



\icmlsetsymbol{equal}{*}

\begin{icmlauthorlist}
\icmlauthor{David Ha}{goo}
\icmlauthor{J\"{u}rgen Schmidhuber}{nnaisense,idsia}
\end{icmlauthorlist}



\icmlaffiliation{goo}{\scriptsize Google Brain}
\icmlaffiliation{nnaisense}{\scriptsize NNAISENSE}
\icmlaffiliation{idsia}{\scriptsize Swiss AI Lab, IDSIA (USI \& SUPSI)}


\icmlkeywords{Machine Learning}

\vskip 0.3in
]


\printAffiliationsAndNotice{}  

\begin{abstract}
We explore building generative neural network models of popular reinforcement learning environments. Our \textit{world model} can be trained quickly in an unsupervised manner to learn a compressed spatial and temporal representation of the environment. By using features extracted from the world model as inputs to an agent, we can train a very compact and simple policy that can solve the required task. We can even train our agent entirely inside of its own hallucinated dream generated by its world model, and transfer this policy back into the actual environment. \\ \\ An interactive version of this paper is available at \url{https://worldmodels.github.io}
\end{abstract}

\section{Introduction}
\label{introduction}

Humans develop a mental model of the world based on what they are able to perceive with their limited senses. The decisions and actions we make are based on this internal model. Jay Wright Forrester, the father of system dynamics, described a mental model as:

\textit{The image of the world around us, which we carry in our head, is just a model. Nobody in his head imagines all the world, government or country. He has only selected concepts, and relationships between them, and uses those to represent the real system.}~\cite{forrester}

\begin{figure}[ht]
\vskip -0.0in
\begin{center}
\centerline{\includegraphics[width=0.9\columnwidth]{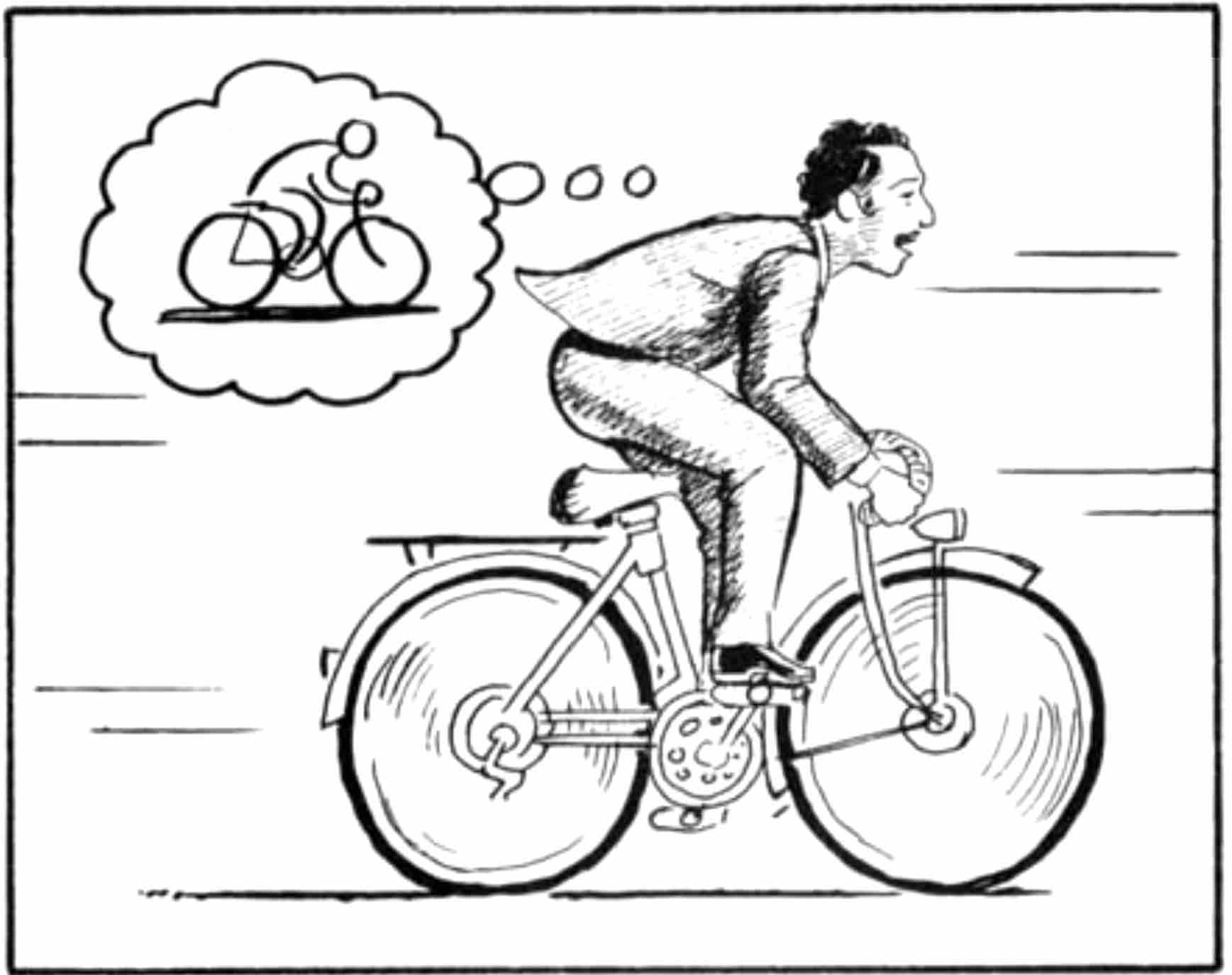}}
\vskip -0.10in
\caption{A World Model, from Scott McCloud's \textit{Understanding Comics.}~\cite{understandingcomics,understandingcomics_blog}}
\label{understanding_comics_figure}
\end{center}
\vskip -0.45in
\end{figure}

To handle the vast amount of information that flows through our daily lives, our brain learns an abstract representation of both spatial and temporal aspects of this information. We are able to observe a scene and remember an abstract description thereof \cite{facial_identity_primate_brain,single_neuron_viz}. Evidence also suggests that what we perceive at any given moment is governed by our brain's prediction of the future based on our internal model \cite{primary_viz_cortex_past_present,mt_motion}.

One way of understanding the predictive model inside of our brains is that it might not be about just predicting the future in general, but predicting future sensory data given our current motor actions \cite{Keller2012,Leinweber2017}. We are able to instinctively act on this predictive model and perform fast reflexive behaviours when we face danger \cite{survival_optimization}, without the need to consciously plan out a course of action.

Take baseball for example. A batter has milliseconds to decide how they should swing the bat -- shorter than the time it takes for visual signals to reach our brain. The reason we are able to hit a 100 mph fastball is due to our ability to instinctively predict when and where the ball will go. For professional players, this all happens subconsciously. Their muscles reflexively swing the bat at the right time and location in line with their internal models' predictions \cite{mt_motion}. They can quickly act on their predictions of the future without the need to consciously roll out possible future scenarios to form a plan \cite{mt_motion_article}.

\begin{figure}[ht]
\vskip -0.0in
\begin{center}
\vskip -0.1in
\centerline{\includegraphics[width=1.0\columnwidth]{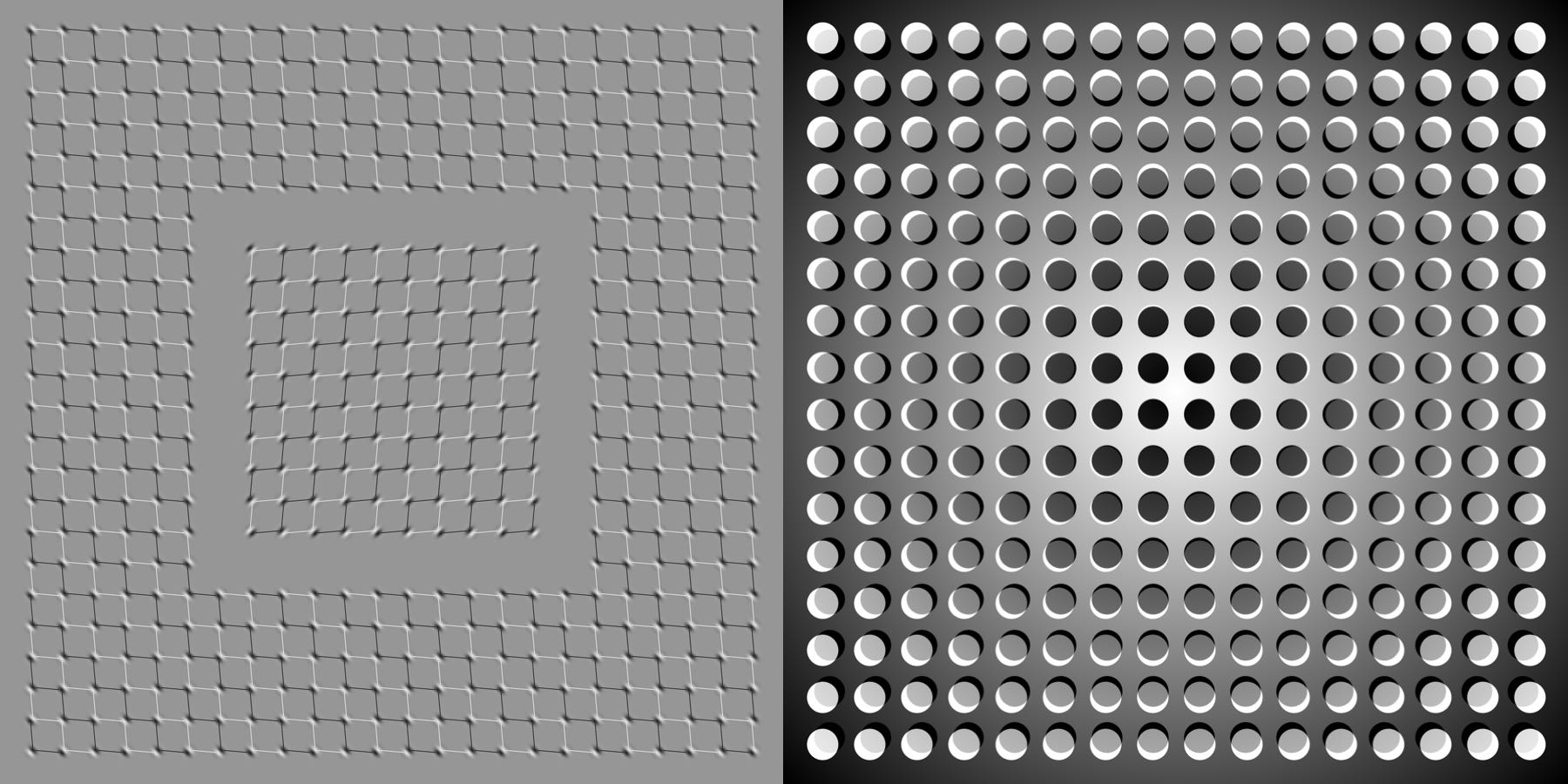}}
\vskip -0.1in
\caption{What we see is based on our brain's prediction of the future \cite{kitaoka,Watanabe2018}.}
\label{kitaoka_illusions_figure}
\end{center}
\vskip -0.3in
\end{figure}

In many reinforcement learning (RL) problems \cite{Kaelbling:96,sutton_barto,wiering2012}, an artificial agent also benefits from having a good representation of past and present states, and a good predictive model of the future \cite{Werbos87specifications,dyna_slides}, preferably a powerful predictive model implemented on a general purpose computer such as a recurrent neural network (RNN) \cite{s05_making_the_world_differentiable,s05a_cm,s05b_rl}.

\begin{figure}[ht]
\vskip -0.0in
\begin{center}
\centerline{\includegraphics[width=1.0\columnwidth]{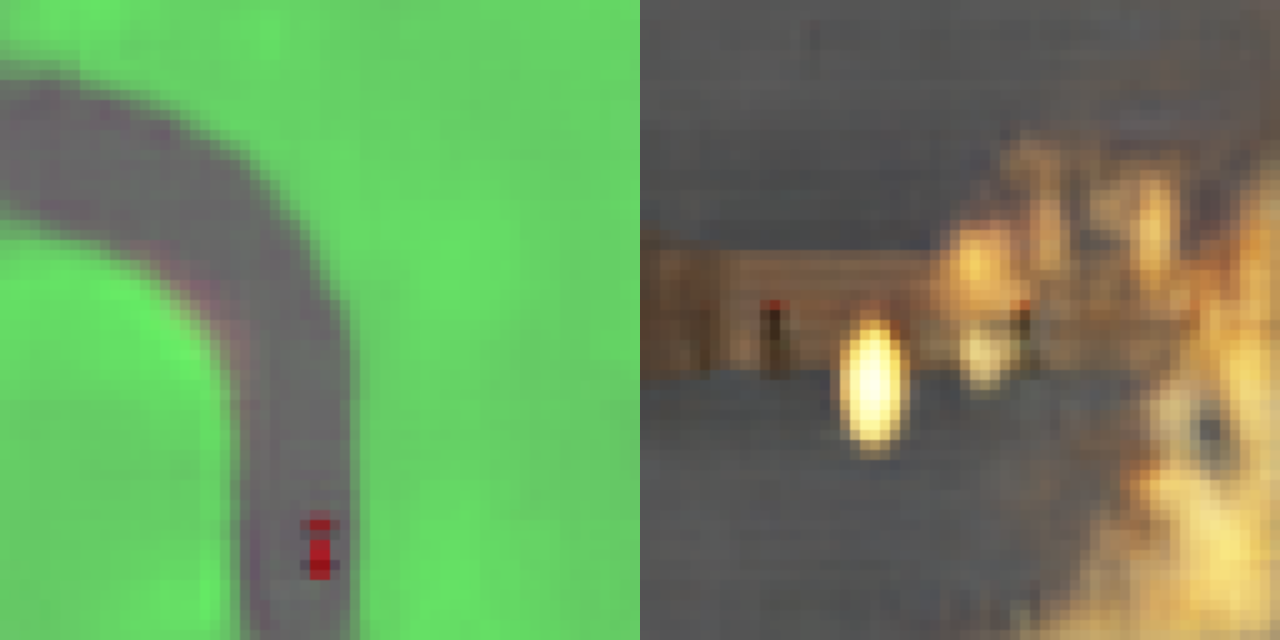}}
\vskip -0.05in
\caption{In this work, we build probabilistic generative models of OpenAI Gym environments. The RNN-based world models are trained using collected observations recorded from the actual game environment. After training the world models, we can use them mimic the complete environment and train agents using them.}
\end{center}
\vskip -0.25in
\end{figure}
Large RNNs are highly expressive models that can learn rich spatial and temporal representations of data. However, many \textit{model-free} RL methods in the literature often only use small neural networks with few parameters. The RL algorithm is often bottlenecked by the credit assignment problem, which makes it hard for traditional RL algorithms to learn millions of weights of a large model, hence in practice, smaller networks are used as they iterate faster to a good policy during training.

Ideally, we would like to be able to efficiently train large RNN-based agents. The backpropagation algorithm \cite{Linnainmaa:1970,Kelley:1960,werbos1982sensitivity} can be used to train large neural networks efficiently. In this work we look at training a large neural network\footnote{Typical model-free RL models have in the order of $10^3$ to $10^6$ model parameters. We look at training models in the order of $10^7$ parameters, which is still rather small compared to state-of-the-art deep learning models with $10^8$ to even $10^{9}$ parameters. In principle, the procedure described in this article can take advantage of these larger networks if we wanted to use them.} to tackle RL tasks, by dividing the agent into a large world model and a small controller model. We first train a large neural network to learn a model of the agent's world in an unsupervised manner, and then train the smaller controller model to learn to perform a task using this world model. A small controller lets the training algorithm focus on the credit assignment problem on a small search space, while not sacrificing capacity and expressiveness via the larger world model. By training the agent through the lens of its world model, we show that it can learn a highly compact policy to perform its task.

Although there is a large body of research relating to \textit{model-based} reinforcement learning, this article is not meant to be a review \cite{rl_survey,s03_overview} of the current state of the field. Instead, the goal of this article is to distill several key concepts from a series of papers 1990--2015 on combinations of RNN-based world models and controllers \cite{s05_making_the_world_differentiable,s05a_cm,s05b_rl,s05c_boredom,learning_to_think}. We will also discuss other related works in the literature that share similar ideas of learning a world model and training an agent using this model.

In this article, we present a simplified framework that we can use to experimentally demonstrate some of the key concepts from these papers, and also suggest further insights to effectively apply these ideas to various RL environments. We use similar terminology and notation as \textit{On Learning to Think: Algorithmic Information Theory for Novel Combinations of RL Controllers and RNN World Models} \cite{learning_to_think} when describing our methodology and experiments.
\vskip -0.15in
\section{Agent Model}
\vskip -0.05in
We present a simple model inspired by our own cognitive system. In this model, our agent has a visual sensory component that compresses what it sees into a small representative code. It also has a memory component that makes predictions about future codes based on historical information. Finally, our agent has a decision-making component that decides what actions to take based only on the representations created by its vision and memory components.

\begin{figure}[ht]
\vskip -0.05in
\begin{center}
\centerline{\includegraphics[width=1.0\columnwidth]{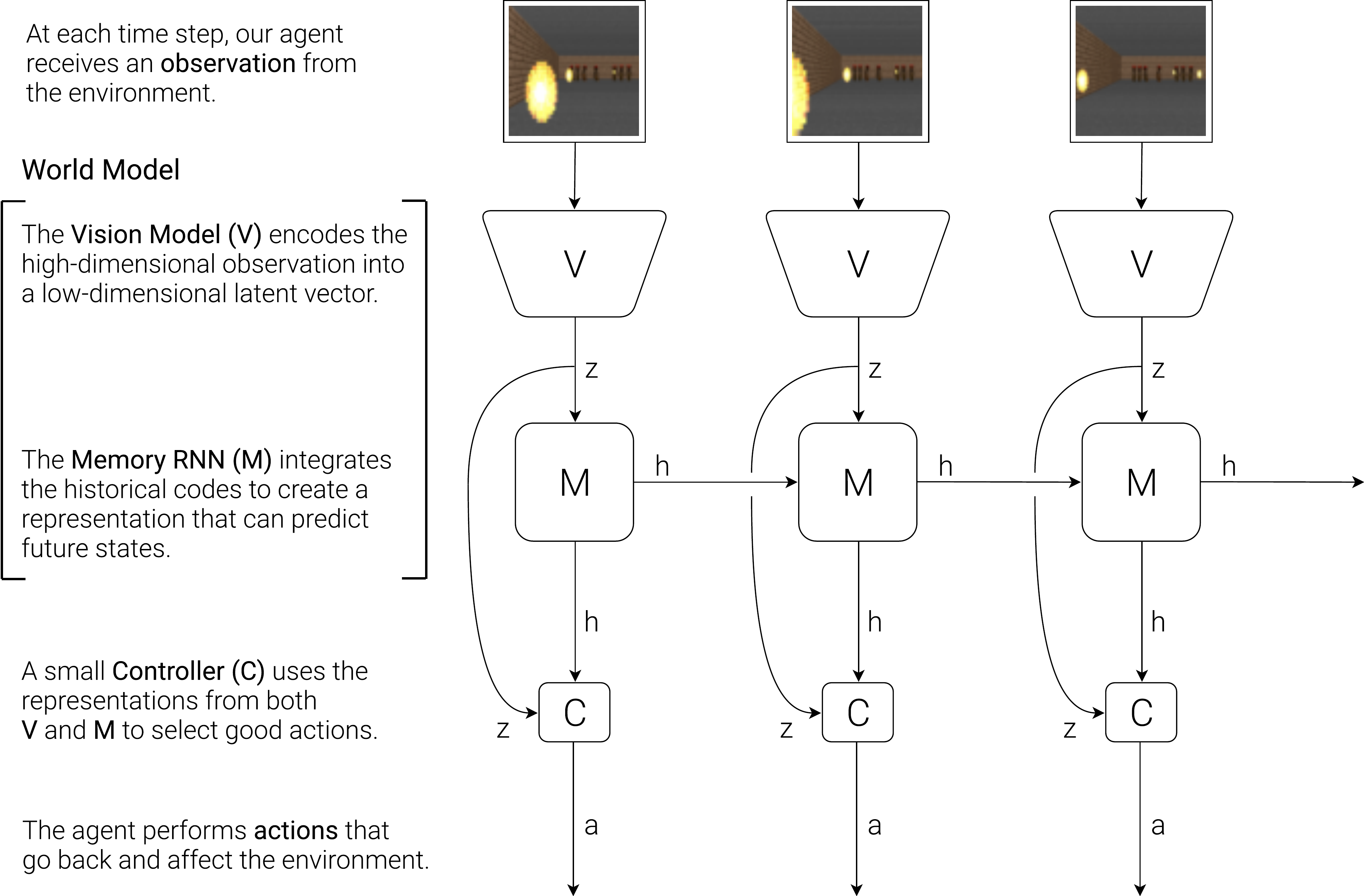}}
\vskip -0.0in
\caption{Our agent consists of three components that work closely together:
\textbf{Vision (V)}, \textbf{Memory (M)}, and \textbf{Controller (C)}}
\label{overview_diagram_figure}
\end{center}
\vskip -0.3in
\end{figure}

\subsection{VAE (V) Model}
\vskip -0.05in
The environment provides our agent with a high dimensional input observation at each time step. This input is usually a 2D image frame that is part of a video sequence. The role of the V model is to learn an abstract, compressed representation of each observed input frame.

\begin{figure}[ht]
\vskip -0.0in
\begin{center}
\centerline{\includegraphics[width=1.0\columnwidth]{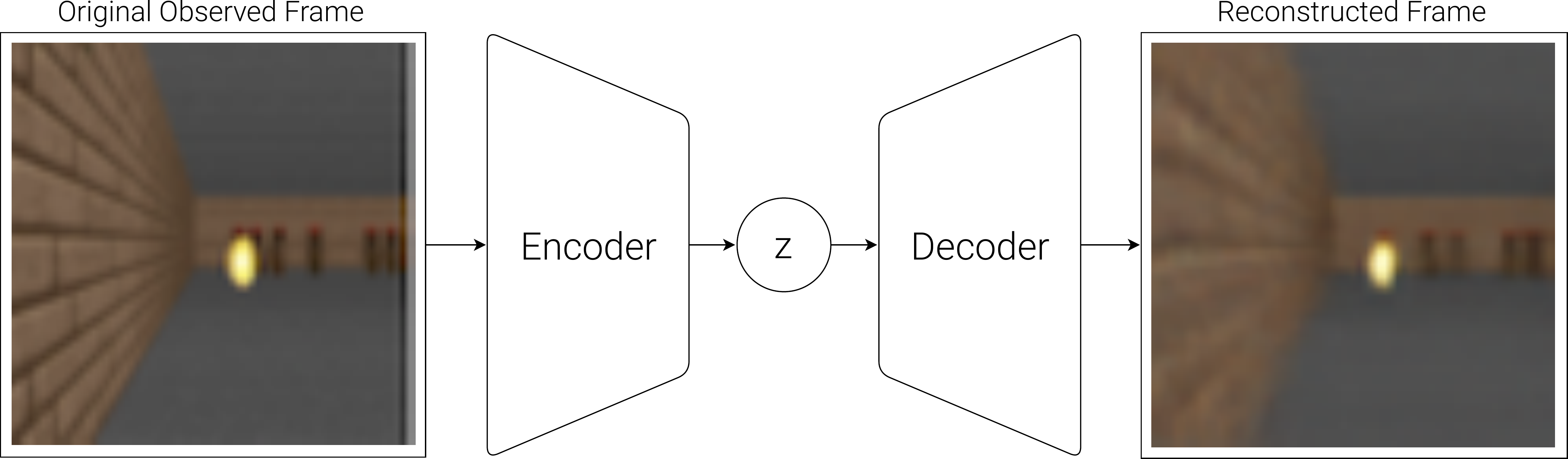}}
\caption{Flow diagram of a Variational Autoencoder (VAE).}
\label{vae_figure}
\end{center}
\vskip -0.3in
\end{figure}

Here, we use a simple Variational Autoencoder~\cite{vae,vae_dm} as our V model to compress each image frame into a small \textit{latent vector} $z$.

\subsection{MDN-RNN (M) Model}

While it is the role of the V model to compress what the agent sees at each time frame, we also want to compress what happens over time. For this purpose, the role of the M model is to predict the future. The M model serves as a predictive model of the future $z$ vectors that V is expected to produce. Since many complex environments are stochastic in nature, we train our RNN to output a probability density function $p(z)$ instead of a deterministic prediction of $z$.

\begin{figure}[ht]
\vskip -0.15in
\begin{center}
\centerline{\includegraphics[width=0.85\columnwidth]{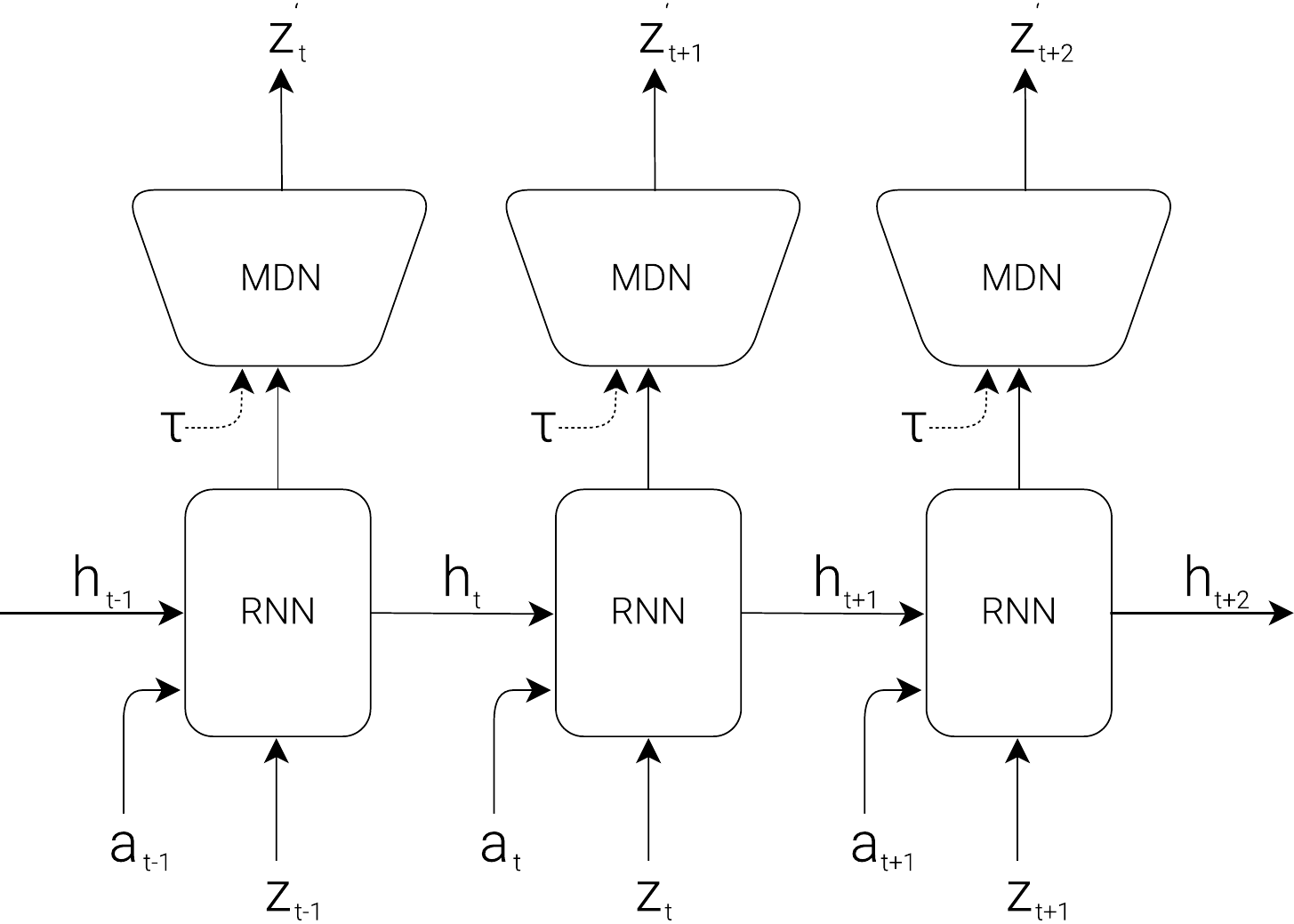}}
\vskip -0.10in
\caption{RNN with a Mixture Density Network output layer. The MDN outputs the parameters of a mixture of Gaussian distribution used to sample a prediction of the next latent vector $z$.}
\label{mdn_rnn_new_figure}
\end{center}
\vskip -0.25in
\end{figure}

In our approach, we approximate $p(z)$ as a mixture of Gaussian distribution, and train the RNN to output the probability distribution of the next latent vector $z_{t+1}$ given the current and past information made available to it.

More specifically, the RNN will model $P(z_{t+1} \; | \; a_t, z_t, h_t)$, where $a_t$ is the action taken at time $t$ and $h_t$ is the \textit{hidden state} of the RNN at time $t$. During sampling, we can adjust a \textit{temperature} parameter $\tau$ to control model uncertainty, as done in \cite{sketchrnn} -- we will find adjusting $\tau$ to be useful for training our controller later on.

\begin{figure}[ht]
\vskip -0.0in
\begin{center}
\centerline{\includegraphics[width=1.0\columnwidth]{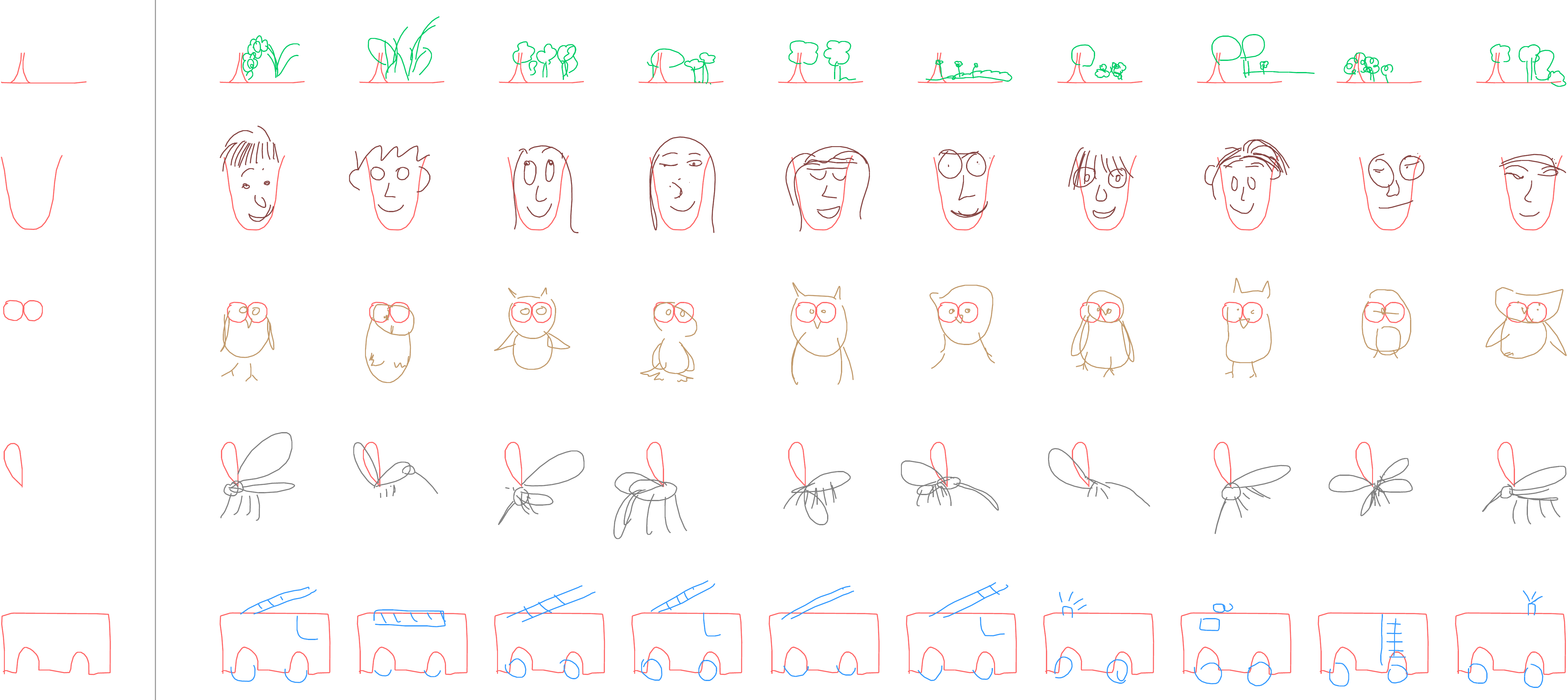}}
\caption{SketchRNN \cite{sketchrnn} is an example of a MDN-RNN used to predict the next pen strokes of a sketch drawing. We use a similar model to predict the next latent vector $z_t$.}
\label{sketchrnn_figure}
\end{center}
\vskip -0.2in
\end{figure}

This approach is known as a Mixture Density Network~\cite{bishop} combined with a RNN (MDN-RNN) \cite{graves_rnn,mdnrnn_tutorial}, and has been applied in the past for sequence generation problems such as generating handwriting \cite{graves_rnn} and sketches \cite{sketchrnn}.

\subsection{Controller (C) Model}

The Controller (C) model is responsible for determining the course of actions to take in order to maximize the expected cumulative reward of the agent during a rollout of the environment. In our experiments, we deliberately make C as simple and small as possible, and trained separately from V and M, so that most of our agent's complexity resides in the world model (V and M).

C is a simple single layer linear model that maps $z_t$ and $h_t$ directly to action $a_t$ at each time step:
\vskip -0.1in
\begin{equation}
\label{controller_equation}
a_t = W_c \; [z_t \; h_t]\; + b_c
\end{equation}
\vskip -0.00in
In this linear model, $W_c$ and $b_c$ are the weight matrix and bias vector that maps the concatenated input vector $[z_t \; h_t]$ to the output action vector $a_t$.

\subsection{Putting V, M, and C Together}

The following flow diagram illustrates how V, M, and C interacts with the environment:

\begin{figure}[ht]
\vskip -0.0in
\begin{center}
\centerline{\includegraphics[width=0.70\columnwidth]{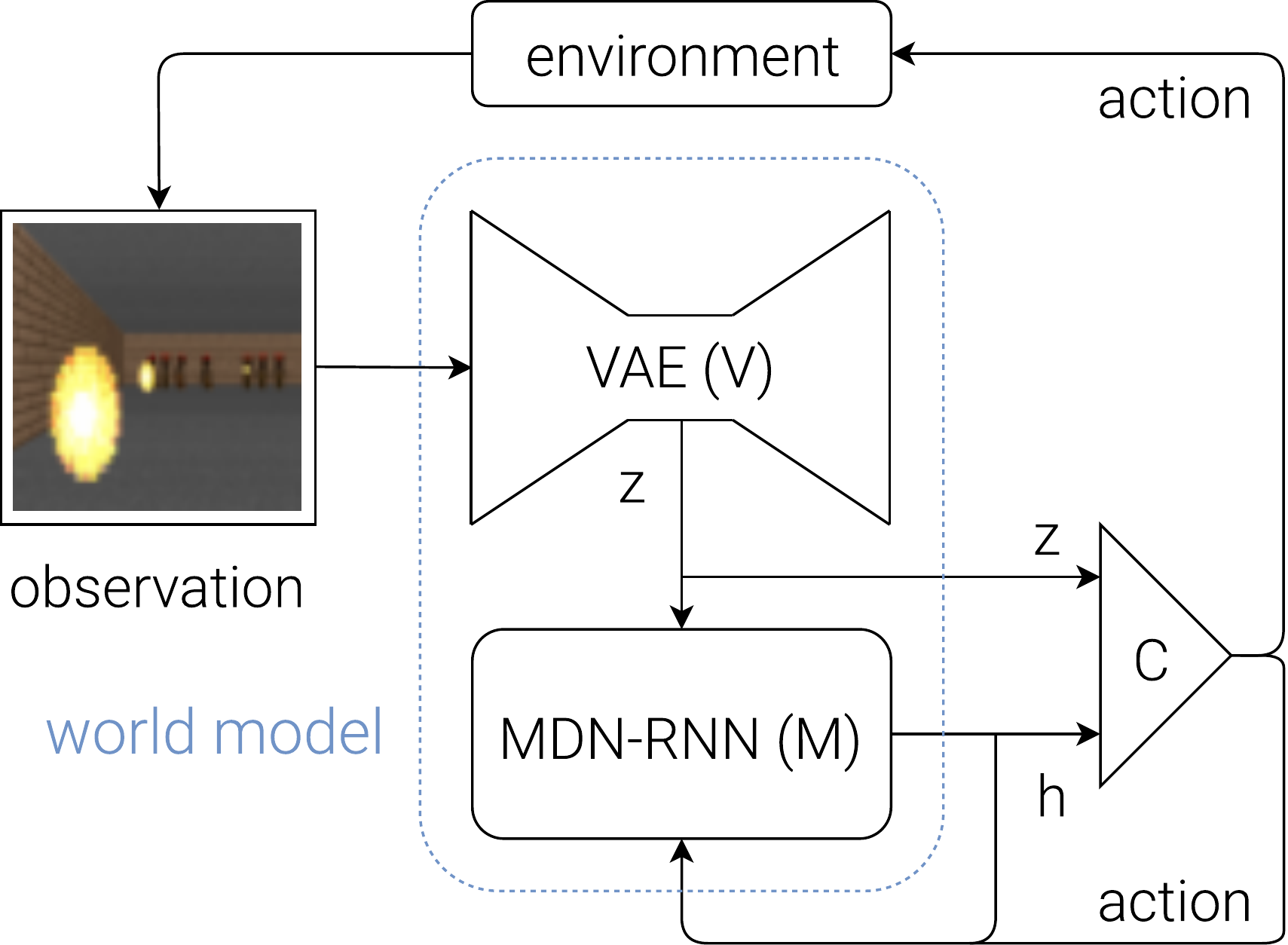}}
\caption{Flow diagram of our Agent model. The raw observation is first processed by V at each time step $t$ to produce $z_t$. The input into C is this latent vector $z_t$ concatenated with M's hidden state $h_t$ at each time step. C will then output an action vector $a_t$ for motor control, and will affect the environment. M will then take the current $z_t$ and action $a_t$ as an input to update its own hidden state to produce $h_{t+1}$ to be used at time $t+1$.}
\label{world_model_schematic_figure}
\end{center}
\vskip -0.3in
\end{figure}

Below is the pseudocode for how our agent model is used in the OpenAI Gym~\cite{openai_gym} environment:
\begin{verbatim}
def rollout(controller):
  ''' env, rnn, vae are '''
  ''' global variables  '''
  obs = env.reset()
  h = rnn.initial_state()
  done = False
  cumulative_reward = 0
  while not done:
    z = vae.encode(obs)
    a = controller.action([z, h])
    obs, reward, done = env.step(a)
    cumulative_reward += reward
    h = rnn.forward([a, z, h])
  return cumulative_reward
\end{verbatim}

Running this function on a given \texttt{controller} C will return the cumulative reward during a rollout.

This minimal design for C also offers important practical benefits. Advances in deep learning provided us with the tools to train large, sophisticated models efficiently, provided we can define a well-behaved, differentiable loss function. Our V and M models are designed to be trained efficiently with the backpropagation algorithm using modern GPU accelerators, so we would like most of the model's complexity, and model parameters to reside in V and M. The number of parameters of C, a linear model, is minimal in comparison. This choice allows us to explore more unconventional ways to train C -- for example, even using evolution strategies (ES)~\cite{Rechenberg1973,Schwefel1977} to tackle more challenging RL tasks where the credit assignment problem is difficult.

To optimize the parameters of C, we chose the Covariance-Matrix Adaptation Evolution Strategy (CMA-ES)~\cite{cmaes,cmaes_original} as our optimization algorithm since it is known to work well for solution spaces of up to a few thousand parameters. We evolve parameters of C on a single machine with multiple CPU cores running multiple rollouts of the environment in parallel.

For more specific information about the models, training procedures, and environments used in our experiments, please refer to the Appendix section.

\section{Car Racing Experiment}

In this section, we describe how we can train the Agent model described earlier to solve a car racing task. To our knowledge, our agent is the first known solution to achieve the score required to solve this task.\footnote{We find this task interesting because although it is not difficult to train an agent to wobble around randomly generated tracks and obtain a mediocre score, \texttt{CarRacing-v0} defines \textit{solving} as getting average reward of 900 over 100 consecutive trials, which means the agent can only afford very few driving mistakes.}

\subsection{World Model for Feature Extraction}

A predictive world model can help us extract useful representations of space and time. By using these features as inputs of a controller, we can train a compact and minimal controller to perform a continuous control task, such as learning to drive from pixel inputs for a top-down car racing environment called  \texttt{CarRacing-v0}~\cite{carracing_v0}.

\begin{figure}[ht]
\vskip -0.0in
\begin{center}
\centerline{\includegraphics[width=1.0\columnwidth]{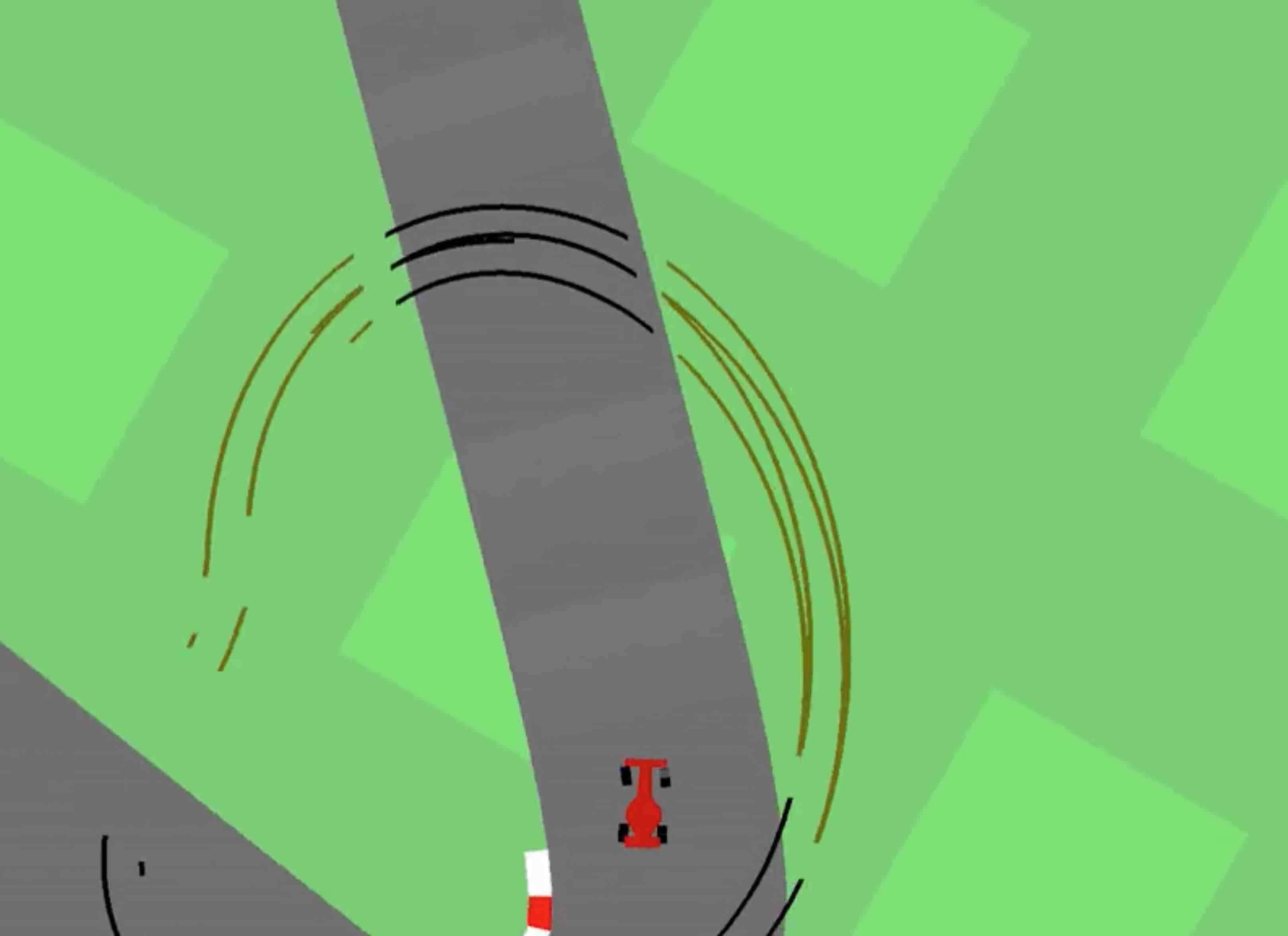}}
\caption{Our agent learning to navigate in \texttt{CarRacing-v0}.}
\end{center}
\vskip -0.3in
\end{figure}

In this environment, the tracks are randomly generated for each trial, and our agent is rewarded for visiting as many tiles as possible in the least amount of time. The agent controls three continuous actions: steering left/right, acceleration, and brake.

To train our V model, we first collect a dataset of 10,000 random rollouts of the environment. We have first an agent acting randomly to explore the environment multiple times, and record the random actions $a_t$ taken and the resulting observations from the environment. We use this dataset to train V to learn a latent space of each frame observed. We train our VAE to encode each frame into low dimensional latent vector $z$ by minimizing the difference between a given frame and the reconstructed version of the frame produced by the decoder from $z$.

We can now use our trained V model to pre-process each frame at time $t$ into $z_t$ to train our M model. Using this pre-processed data, along with the recorded random actions $a_t$ taken, our MDN-RNN can now be trained to model $P(z_{t+1} \; | \; a_t, z_t, h_t)$ as a mixture of Gaussians.\footnote{In principle, we can train both models together in an end-to-end manner, although we found that training each separately is more practical, and also achieves satisfactory results. Training each model only required less than an hour of computation time on a single GPU. We can also train individual VAE and MDN-RNN models without having to exhaustively tune hyperparameters.}

In this experiment, the world model (V and M) has no knowledge about the actual reward signals from the environment. Its task is simply to compress and predict the sequence of image frames observed. Only the Controller (C) Model has access to the reward information from the environment. Since there are a mere 867 parameters inside the linear controller model, evolutionary algorithms such as CMA-ES are well suited for this optimization task.

We can use the VAE to reconstruct each frame using $z_t$ at each time step to visualize the quality of the information the agent actually sees during a rollout. The figure below is a VAE model trained on screenshots from \texttt{CarRacing-v0}.

\begin{figure}[ht]
\vskip -0.0in
\begin{center}
\centerline{\includegraphics[width=1.0\columnwidth]{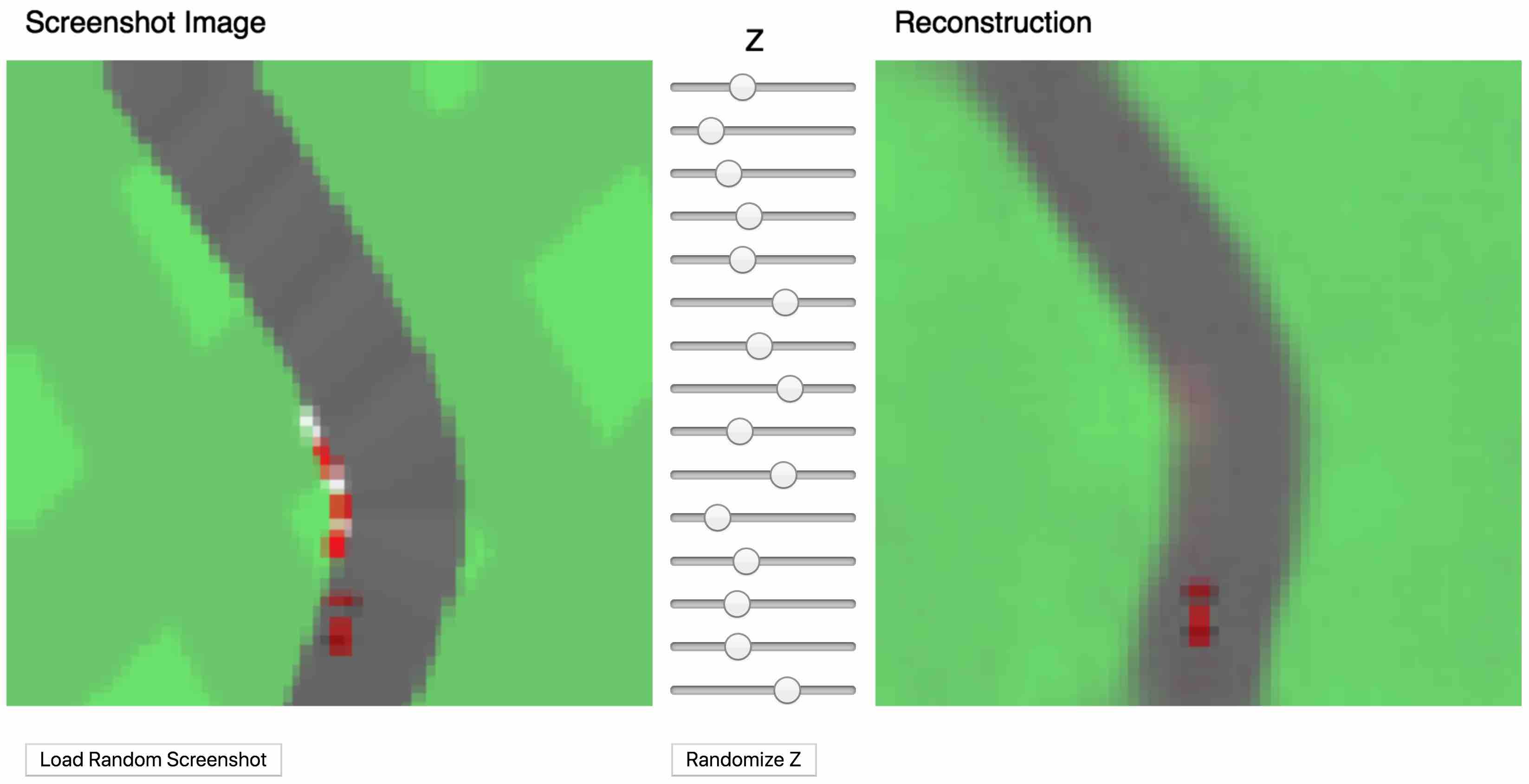}}
\caption{Despite losing details during this lossy compression process, latent vector $z$ captures the essence of each image frame.}
\end{center}
\vskip -0.3in
\end{figure}

In the online version of this article, one can load randomly chosen screenshots to be encoded into a small latent vector $z$, which is used to reconstruct the original screenshot. One can also experiment with adjusting the values of the $z$ vector using the slider bars to see how it affects the reconstruction, or randomize $z$ to observe the space of possible screenshots.

\subsection{Procedure}

To summarize the Car Racing experiment, below are the steps taken:

\begin{enumerate}
  \item Collect 10,000 rollouts from a random policy.
  \item  Train VAE (V) to encode frames into $z \in \mathcal{R}^{32}$.
  \item  Train MDN-RNN (M) to model $P(z_{t+1} \; | \; a_t, z_t, h_t)$.
  \item  Define Controller (C) as $a_t = W_c \; [z_t \; h_t]\; + \; b_c$.
  \item  Use CMA-ES to solve for a $W_c$ and $b_c$ that maximizes the expected cumulative reward.
\end{enumerate}

\begin{table}[ht]
\label{car_racing_param_count_table}
\vskip -0.0in
\begin{center}
\begin{small}
\begin{sc}
\begin{tabular}{lc}
\toprule
Model & Parameter Count  \\
\midrule
VAE    & 4,348,547 \\
MDN-RNN & 422,368 \\
Controller    & 867 \\
\bottomrule
\end{tabular}
\end{sc}
\end{small}
\end{center}
\vskip -0.0in
\end{table}

\subsection{Experiment Results}
\medskip
\textit{V Model Only}

Training an agent to drive is not a difficult task if we have a good representation of the observation. Previous works~\cite{browser_car,mar_io_kart,keras_car} have shown that with a good set of hand-engineered information about the observation, such as LIDAR information, angles, positions and velocities, one can easily train a small feed-forward network to take this hand-engineered input and output a satisfactory navigation policy. For this reason, we first want to test our agent by handicapping C to only have access to V but not M, so we define our controller as $a_t = W_c \; z_t \;+ \; b_c$.

\begin{figure}[ht]
\vskip -0.0in
\begin{center}
\centerline{\includegraphics[width=1.0\columnwidth]{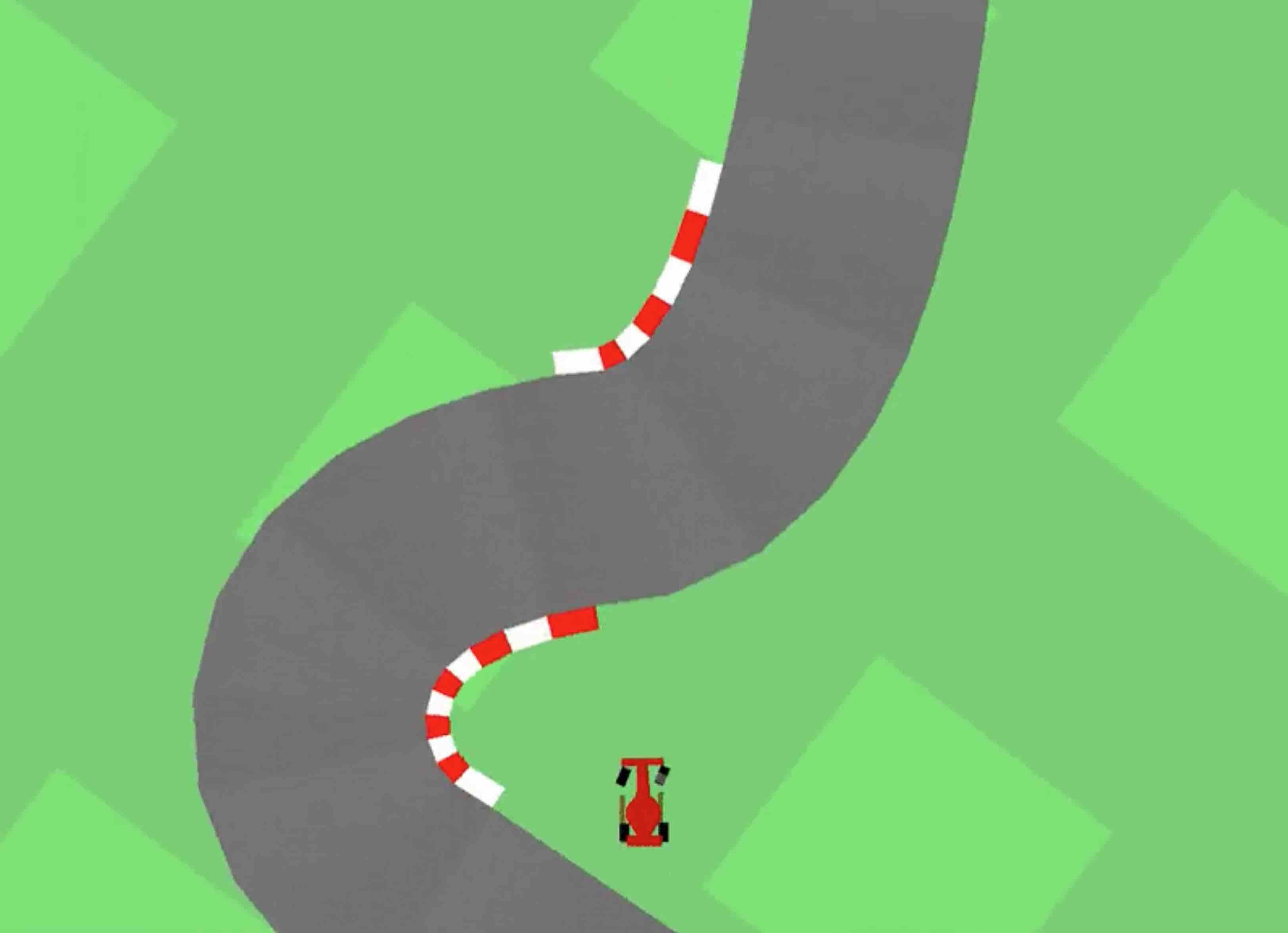}}
\caption{Limiting our controller to see only $z_t$, but not $h_t$ results in wobbly and unstable driving behaviours.}
\end{center}
\vskip -0.3in
\end{figure}

Although the agent is still able to navigate the race track in this setting, we notice it wobbles around and misses the tracks on sharper corners. This handicapped agent achieved an average score of 632 $\pm$ 251 over 100 random trials, in line with the performance of other agents on OpenAI Gym's leaderboard~\cite{carracing_v0} and traditional Deep RL methods such as A3C~\cite{carracing_cs221,carracing_cs234}. Adding a hidden layer to C's policy network helps to improve the results to 788 $\pm$ 141, but not quite enough to solve this environment.

\medskip

\textit{Full World Model (V and M)}

The representation $z_t$ provided by our V model only captures a representation at a moment in time and does not have much predictive power. In contrast, M is trained to do one thing, and to do it really well, which is to predict $z_{t+1}$. Since M's prediction of $z_{t+1}$ is produced from the RNN's hidden state $h_t$ at time $t$, this vector is a good candidate for the set of learned features we can give to our agent. Combining $z_t$ with $h_t$ gives our controller C a good representation of both the current observation, and what to expect in the future.

\begin{figure}[ht]
\vskip -0.0in
\begin{center}
\centerline{\includegraphics[width=1.0\columnwidth]{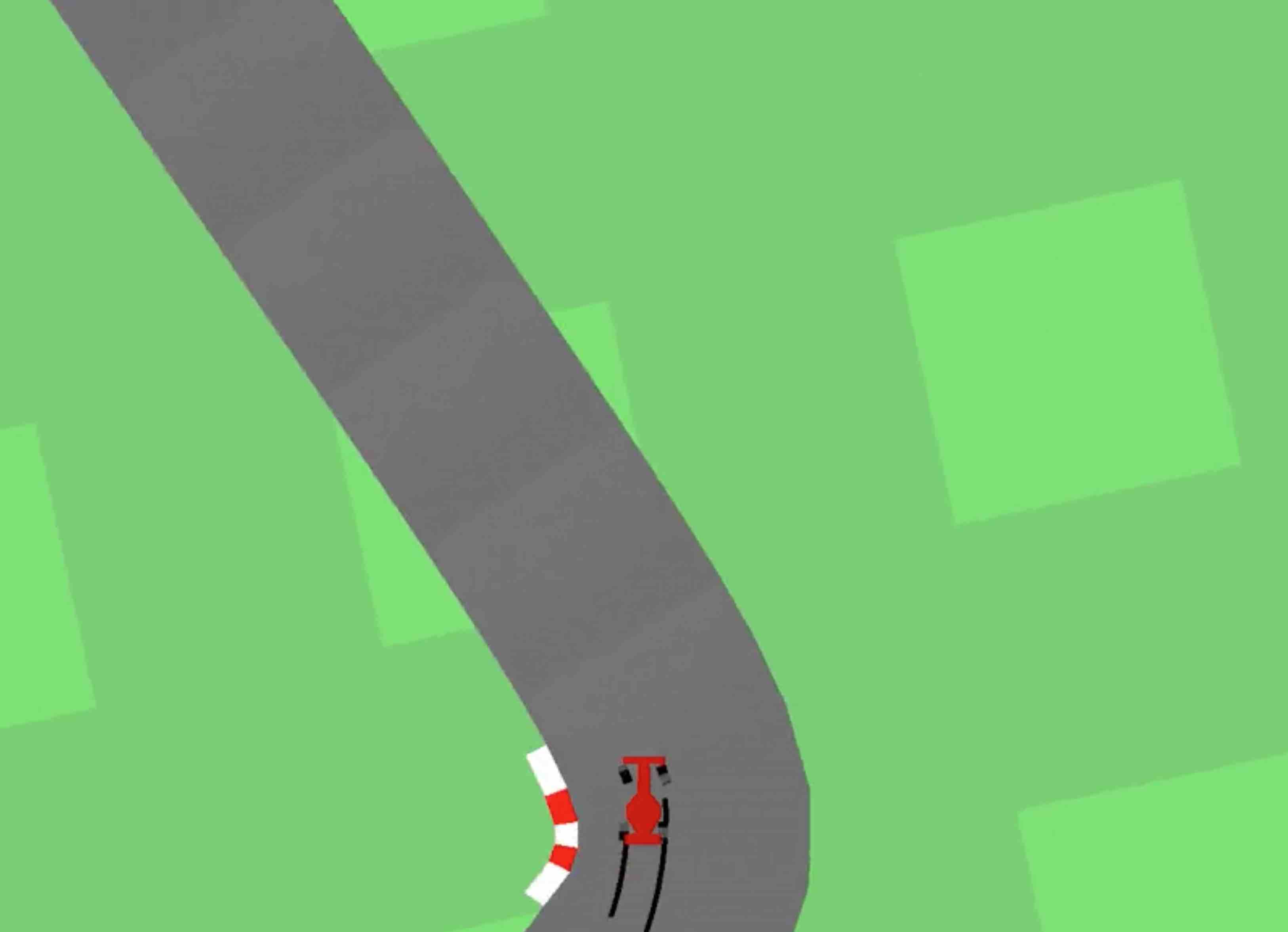}}
\caption{Driving is more stable if we give our controller access to both $z_t$ and $h_t$.}
\end{center}
\vskip -0.4in
\end{figure}

We see that allowing the agent to access the both $z_t$ and $h_t$ greatly improves its driving capability. The driving is more stable, and the agent is able to seemingly attack the sharp corners effectively. Furthermore, we see that in making these fast reflexive driving decisions during a car race, the agent does not need to \textit{plan ahead} and roll out hypothetical scenarios of the future. Since $h_t$ contain information about the probability distribution of the future, the agent can just query the RNN instinctively to guide its action decisions. Like a seasoned Formula One driver or the baseball player discussed earlier, the agent can instinctively predict when and where to navigate in the heat of the moment.

\begin{table}[ht]
\label{car_racing_table}
\vskip -0.0in
\begin{center}
\begin{small}
\begin{sc}
\begin{tabular}{ll}
\toprule
Method & Avg. Score \\
\midrule
DQN~\cite{dqn_racecar} & 343 $\pm$ 18 \\
A3C (continuous)~\cite{carracing_cs234} & 591 $\pm$ 45 \\
A3C (discrete)~\cite{carracing_cs221} & 652 $\pm$ 10 \\
ceobillionaire (Gym Leaderboard) & 838 $\pm$ 11 \\
V model & 632 $\pm$ 251 \\
V model with hidden layer & 788 $\pm$ 141 \\
\textbf{Full World Model} & \textbf{906 $\pm$ 21} \\
\bottomrule
\end{tabular}
\end{sc}
\end{small}
\end{center}
\vskip -0.2in
\caption{\texttt{CarRacing-v0} scores achieved using various methods.}
\vskip -0.1in
\end{table}

Our agent is able to achieve a score of 906 $\pm$ 21 over 100 random trials, effectively solving the task and obtaining new state of the art results. Previous attempts~\cite{carracing_cs221,carracing_cs234} using Deep RL methods obtained average scores of 591--652 range, and the best reported solution on the leaderboard obtained an average score of 838 $\pm$ 11 over 100 random trials. Traditional Deep RL methods often require pre-processing of each frame, such as employing edge-detection~\cite{carracing_cs234}, in addition to stacking a few recent frames~\cite{carracing_cs221,carracing_cs234} into the input. In contrast, our world model takes in a stream of raw RGB pixel images and directly learns a spatial-temporal representation. To our knowledge, our method is the first reported solution to solve this task.

\subsection{Car Racing Dreams}

Since our world model is able to model the future, we are also able to have it come up with hypothetical car racing scenarios on its own. We can ask it to produce the probability distribution of $z_{t+1}$ given the current states, \textit{sample} a $z_{t+1}$ and use this sample as the real observation. We can put our trained C back into this hallucinated environment generated by M. The following image from an interactive demo in the online version of this article shows how our world model can be used to hallucinate the car racing environment:

\begin{figure}[ht]
\vskip -0.0in
\begin{center}
\centerline{\includegraphics[width=1.0\columnwidth]{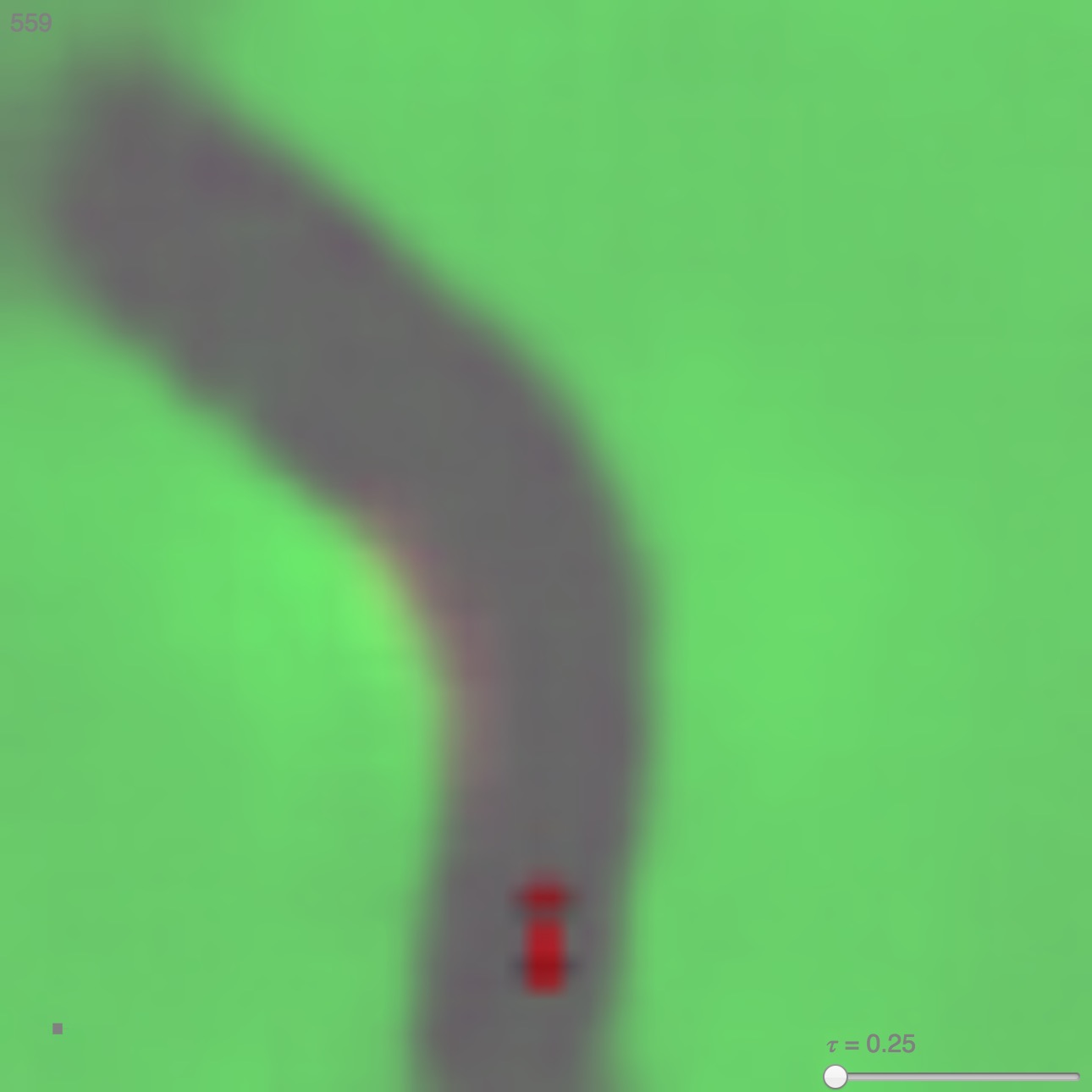}}
\caption{Our agent driving inside of its own dream world. Here, we deploy our trained policy into a fake environment generated by the MDN-RNN, and rendered using the VAE's decoder. In the demo, one can override the agent's actions as well as adjust $\tau$ to control the uncertainty of the environment generated by M.}
\end{center}
\vskip -0.4in
\end{figure}

\section{VizDoom Experiment}

\subsection{Learning Inside of a Dream}

We have just seen that a policy learned inside of the real environment appears to somewhat function inside of the dream environment. This begs the question -- can we train our agent to learn inside of its own dream, and transfer this policy back to the actual environment?

If our world model is sufficiently accurate for its purpose, and complete enough for the problem at hand, we should be able to substitute the actual environment with this world model. After all, our agent does not directly observe the reality, but only sees what the world model lets it see. In this experiment, we train an agent inside the hallucination generated by its world model trained to mimic a VizDoom~\cite{vizdoom} environment.

\begin{figure}[ht]
\vskip -0.0in
\begin{center}
\centerline{\includegraphics[width=1.0\columnwidth]{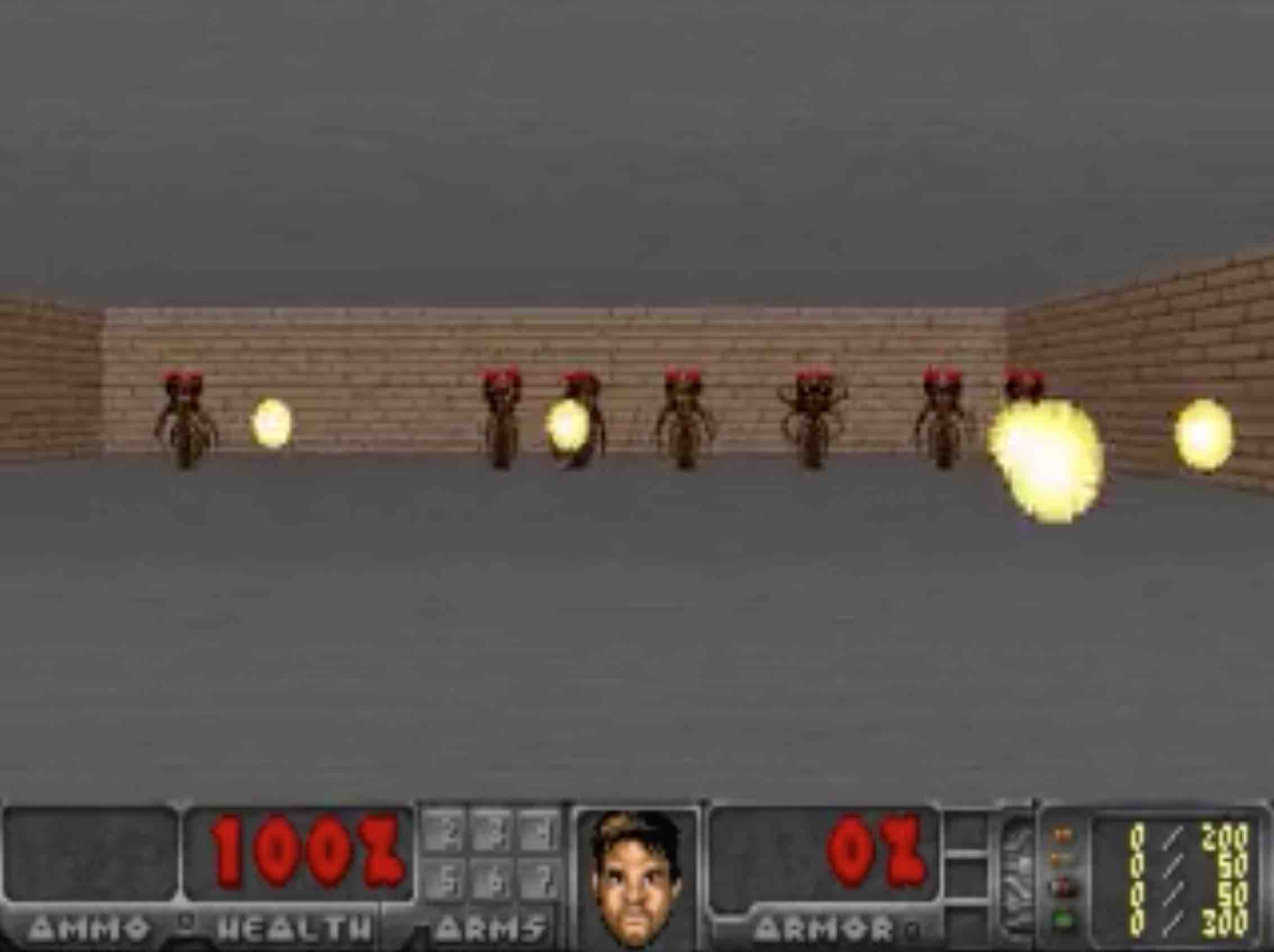}}
\caption{Our final agent solving \textit{VizDoom: Take Cover}.}
\end{center}
\vskip -0.4in
\end{figure}

The agent must learn to avoid fireballs shot by monsters from the other side of the room with the sole intent of killing the agent. There are no explicit rewards in this environment, so to mimic natural selection, the cumulative reward can be defined to be the number of time steps the agent manages to stay alive during a rollout. Each rollout of the environment runs for a maximum of 2100 time steps ($\sim$ 60 seconds), and the task is considered solved if the average survival time over 100 consecutive rollouts is greater than 750 time steps ($\sim$ 20 seconds)~\cite{takecover}.

\subsection{Procedure}

The setup of our VizDoom experiment is largely the same as the Car Racing task, except for a few key differences. In the Car Racing task, M is only trained to model the next $z_{t}$. Since we want to build a world model we can train our agent in, our M model here will also predict whether the agent dies in the next frame (as a binary event $done_t$, or $d_t$ for short), in addition to the next frame $z_t$.

Since the M model can predict the $done$ state in addition to the next observation, we now have all of the ingredients needed to make a full RL environment. We first build an OpenAI Gym environment interface by wrapping a \texttt{gym.Env} interface over our M if it were a real Gym environment, and then train our agent inside of this \textit{virtual} environment instead of using the actual environment.

In this simulation, we do not need the V model to encode any real pixel frames during the hallucination process, so our agent will therefore only train entirely in a latent space environment. This has many advantages as we will see.

This virtual environment has an identical interface to the real environment, so after the agent learns a satisfactory policy in the virtual environment, we can easily deploy this policy back into the actual environment to see how well the policy transfers over.

To summarize the \textit{Take Cover} experiment, below are the steps taken:

\begin{enumerate}
  \item Collect 10,000 rollouts from a random policy.
  \item Train VAE (V) to encode each frame into a latent vector $z \in \mathcal{R}^{64}$, and use V to convert the images collected from (1) into the latent space representation.
  \item  Train MDN-RNN (M) to model \\ $P(z_{t+1}, d_{t+1} \; | \; a_t, z_t, h_t)$.
  \item  Define Controller (C) as $a_t = W_c \; [z_t \; h_t]$.
  \item  Use CMA-ES to solve for a $W_c$ that maximizes the expected survival time inside the virtual environment.
  \item Use learned policy from (5) on actual environment.
\end{enumerate}

\begin{table}[ht]
\label{doom_cover_table}
\vskip -0.0in
\begin{center}
\begin{small}
\begin{sc}
\begin{tabular}{lc}
\toprule
Model & Parameter Count  \\
\midrule
VAE    & 4,446,915 \\
MDN-RNN & 1,678,785 \\
Controller    & 1,088 \\
\bottomrule
\end{tabular}
\end{sc}
\end{small}
\end{center}
\vskip -0.0in
\end{table}

\subsection{Training Inside of the Dream}

After some training, our controller learns to navigate around the dream environment and escape from deadly fireballs launched by monsters generated by M. Our agent achieved a \textit{score} of $\sim$ 900 time steps in the virtual environment.

\begin{figure}[ht]
\vskip -0.0in
\begin{center}
\centerline{\includegraphics[width=1.0\columnwidth]{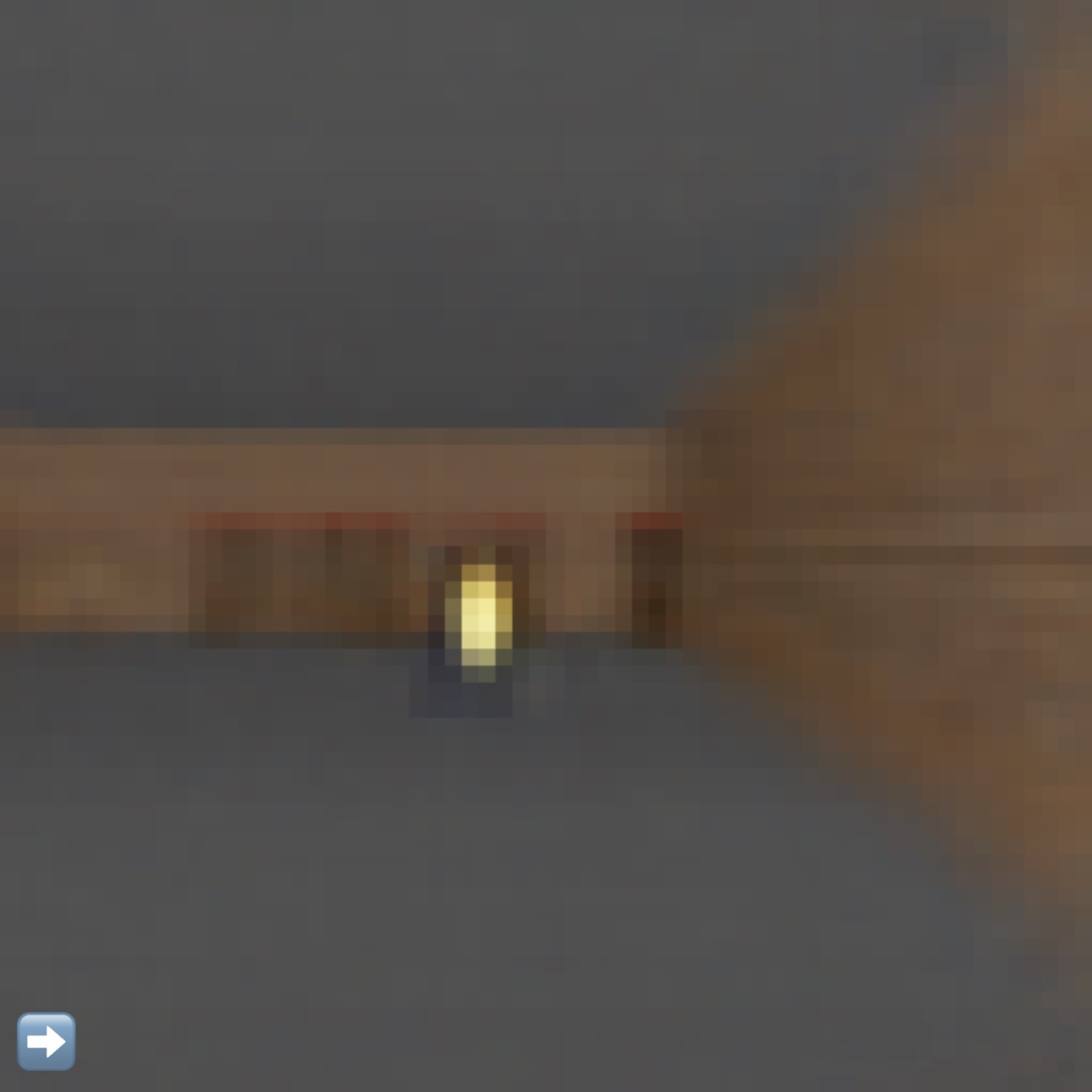}}
\vskip -0.1in
\caption{Our agent discovers a policy to avoid hallucinated fireballs. In the online version of this article, the reader can interact with the environment inside this demo.}
\end{center}
\vskip -0.4in
\end{figure}

Here, our RNN-based world model is trained to mimic a complete game environment designed by human programmers. By learning only from raw image data collected from random episodes, it learns how to simulate the essential aspects of the game -- such as the game logic, enemy behaviour, physics, and also the 3D graphics rendering.

For instance, if the agent selects the left action, the M model learns to move the agent to the left and adjust its internal representation of the game states accordingly. It also learns to block the agent from moving beyond the walls on both sides of the level if the agent attempts to move too far in either direction. Occasionally, the M model needs to keep track of multiple fireballs being shot from several different monsters and coherently move them along in their intended directions. It must also detect whether the agent has been killed by one of these fireballs.

Unlike the actual game environment, however, we note that it is possible to add extra uncertainty into the virtual environment, thus making the game more challenging in the dream environment. We can do this by increasing the temperature $\tau$ parameter during the sampling process of $z_{t+1}$. By increasing the uncertainty, our dream environment becomes more difficult compared to the actual environment. The fireballs may move more randomly in a less predictable path compared to the actual game. Sometimes the agent may even die due to sheer misfortune, without explanation.

We find agents that perform well in higher temperature settings generally perform better in the normal setting. In fact, increasing $\tau$ helps prevent our controller from taking advantage of the imperfections of our world model -- we will discuss this in more depth later on.

\subsection{Transfer Policy to Actual Environment}

\begin{figure}[ht]
\vskip -0.0in
\begin{center}
\centerline{\includegraphics[width=1.0\columnwidth]{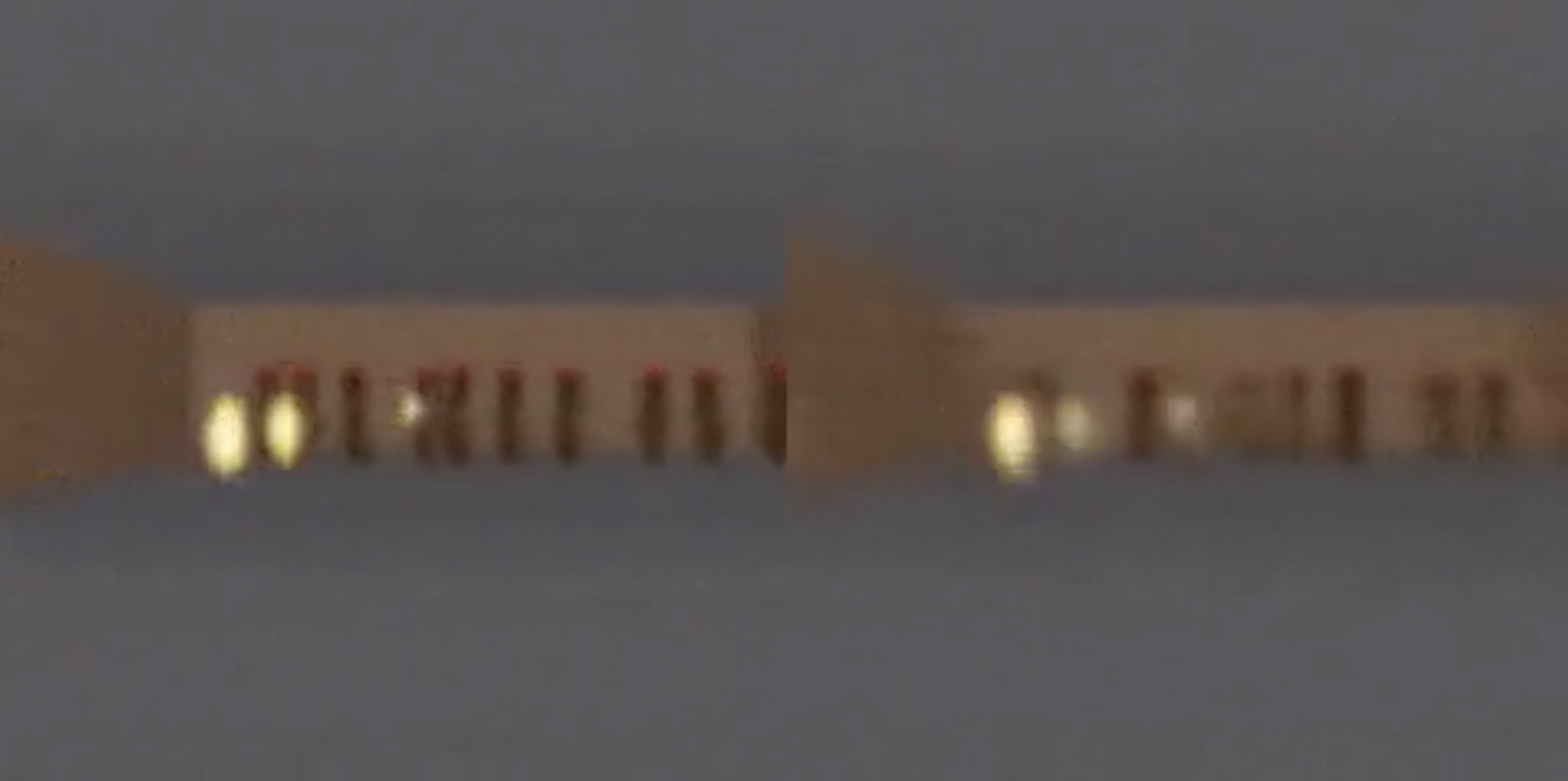}}
\vskip -0.1in
\caption{Deploying our policy learned inside of the dream RNN environment back into the actual VizDoom environment.}
\end{center}
\vskip -0.2in
\end{figure}

We took the agent trained inside of the virtual environment and tested its performance on the original VizDoom scenario. The score over 100 random consecutive trials is $\sim$ 1100 time steps, far beyond the required score of 750 time steps, and also much higher than the score obtained inside the more difficult virtual environment.

\begin{figure}[ht]
\vskip -0.0in
\begin{center}
\centerline{\includegraphics[width=1.0\columnwidth]{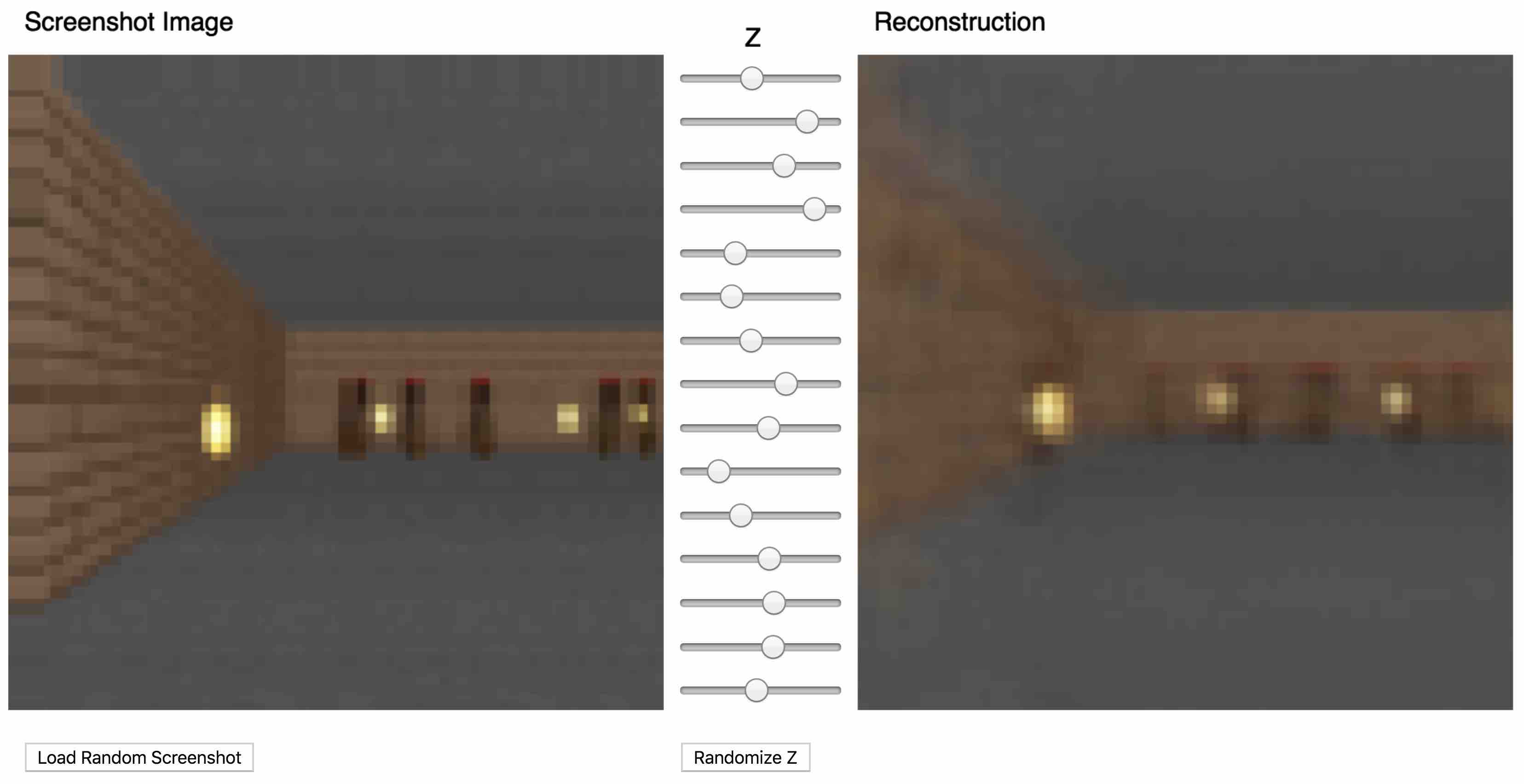}}
\caption{An interactive VAE of Doom in the online article.}
\vskip -0.15in
\end{center}
\vskip -0.25in
\end{figure}

We see that even though the V model is not able to capture all of the details of each frame correctly, for instance, getting the number of monsters correct, the agent is still able to use the learned policy to navigate in the real environment. As the virtual environment cannot even keep track of the exact number of monsters in the first place, an agent that is able to survive the noisier and uncertain virtual nightmare environment will thrive in the original, cleaner environment.

\subsection{Cheating the World Model}

In our childhood, we may have encountered ways to exploit video games in ways that were not intended by the original game designer \cite{video_game_exploits}. Players discover ways to collect unlimited lives or health, and by taking advantage of these exploits, they can easily complete an otherwise difficult game. However, in the process of doing so, they may have forfeited the opportunity to learn the skill required to master the game as intended by the game designer.

For instance, in our initial experiments, we noticed that our agent discovered an \textit{adversarial} policy to move around in such a way so that the monsters in this virtual environment governed by the M model never shoots a single fireball during some rollouts. Even when there are signs of a fireball forming, the agent will move in a way to extinguish the fireballs magically as if it has superpowers in the environment.

Because our world model is only an approximate probabilistic model of the environment, it will occasionally generate trajectories that do not follow the laws governing the actual environment. As we saw previously, even the number of monsters on the other side of the room in the actual environment is not exactly reproduced by the world model. Like a child who learns that objects in the air usually fall to the ground, the child might also imagine unrealistic superheroes who fly across the sky. For this reason, our world model will be exploitable by the controller, even if in the actual environment such exploits do not exist.

And since we are using the M model to generate a virtual dream environment for our agent, we are also giving the controller access to all of the hidden states of M. This is essentially granting our agent access to all of the internal states and memory of the game engine, rather than only the game observations that the player gets to see. Therefore our agent can efficiently explore ways to directly manipulate the hidden states of the game engine in its quest to maximize its expected cumulative reward. The weakness of this approach of learning a policy inside a learned dynamics model is that our agent can easily find an adversarial policy that can fool our dynamics model -- it'll find a policy that looks good under our dynamics model, but will fail in the actual environment, usually because it visits states where the model is wrong because they are away from the training distribution.

\begin{figure}[ht]
\vskip -0.0in
\begin{center}
\centerline{\includegraphics[width=1.0\columnwidth]{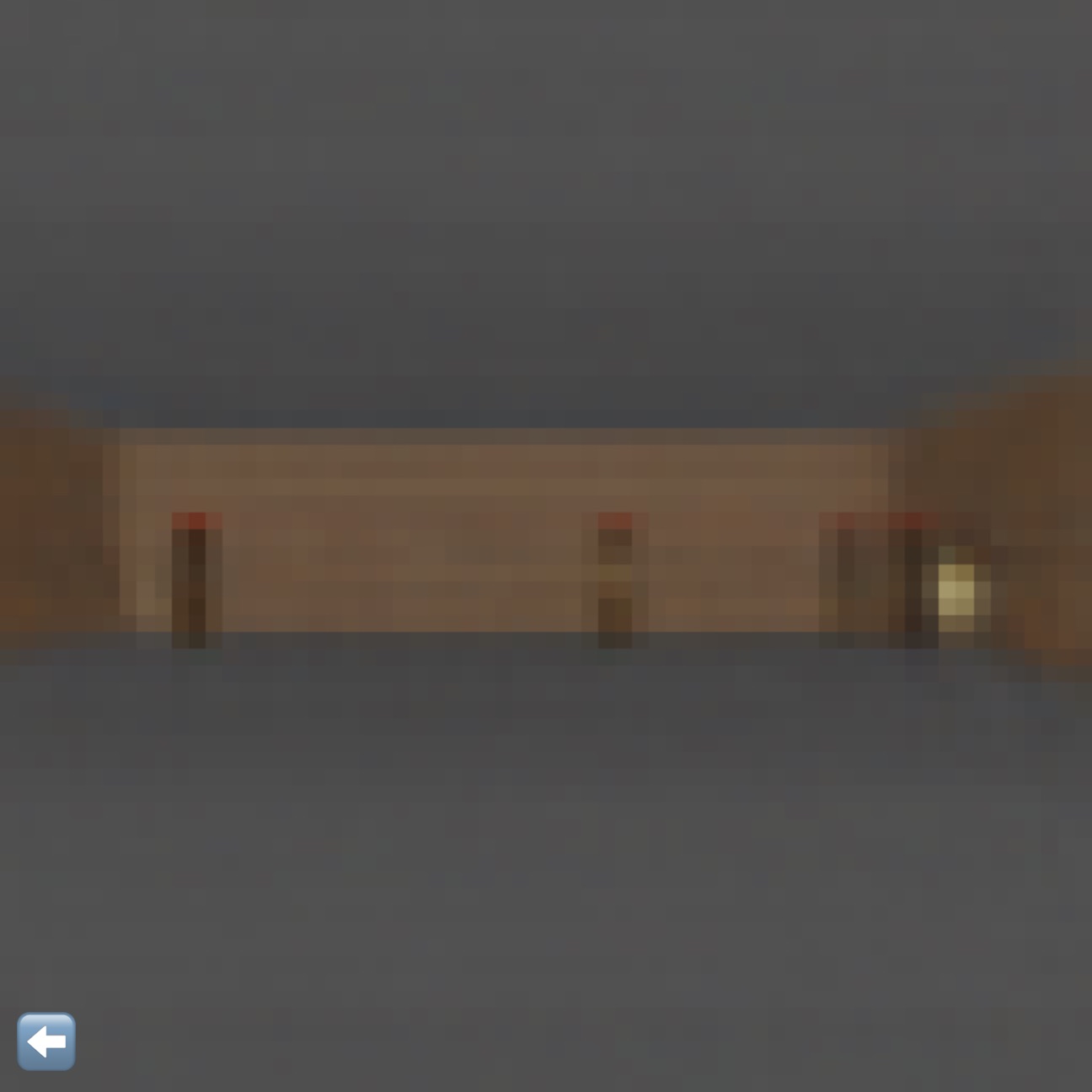}}
\vskip -0.05in
\caption{Agent discovers an adversarial policy to automatically extinguish fireballs after they are fired during some rollouts.}
\end{center}
\vskip -0.20in
\end{figure}

This weakness could be the reason that many previous works that learn dynamics models of RL environments but do not actually use those models to fully replace the actual environments \cite{action_conditional_video_prediction,recurrent_env_sim}. Like in the M model proposed in \cite{s05_making_the_world_differentiable,s05a_cm,s05b_rl}, the dynamics model is a deterministic model, making the model easily exploitable by the agent if it is not perfect. Using Bayesian models, as in PILCO~\cite{pilco}, helps to address this issue with the uncertainty estimates to some extent, however, they do not fully solve the problem. Recent work~\cite{Nagabandi2017} combines the model-based approach with traditional model-free RL training by first initializing the policy network with the learned policy, but must subsequently rely on model-free methods to fine-tune this policy in the actual environment.

In \textit{Learning to Think}~\cite{learning_to_think}, it is acceptable that the RNN M is not always a reliable predictor. A (potentially evolution-based) RNN C can in principle learn to ignore a flawed M, or exploit certain useful parts of M for arbitrary computational purposes including hierarchical planning etc. This is not what we do here though -- our present approach is still closer to some of the older systems \cite{s05_making_the_world_differentiable,s05a_cm,s05b_rl}, where a RNN M is used to predict and plan ahead step by step. Unlike this early work, however, we use evolution for C (like in \textit{Learning to Think}) rather than traditional RL combined with RNNs, which has the advantage of both simplicity and generality.

To make it more difficult for our C model to exploit deficiencies of the M model, we chose to use the MDN-RNN as the dynamics model, which models the \textit{distribution} of possible outcomes in the actual environment, rather than merely predicting a deterministic future. Even if the actual environment is deterministic, the MDN-RNN would in effect approximate it as a stochastic environment. This has the advantage of allowing us to train our C model inside a more stochastic version of any environment -- we can simply adjust the temperature parameter $\tau$ to control the amount of randomness in the M model, hence controlling the tradeoff between realism and exploitability.

Using a mixture of Gaussian model may seem like overkill given that the latent space encoded with the VAE model is just a single diagonal Gaussian distribution. However, the discrete modes in a mixture density model is useful for environments with random discrete events, such as whether a monster decides to shoot a fireball or stay put. While a single diagonal Gaussian might be sufficient to encode individual frames, a RNN with a mixture density output layer makes it easier to model the logic behind a more complicated environment with discrete random states.

For instance, if we set the temperature parameter to a very low value of $\tau=0.1$, effectively training our C model with a M model that is almost identical to a deterministic LSTM, the monsters inside this dream environment fail to shoot fireballs, no matter what the agent does, due to mode collapse. The M model is not able to \textit{jump} to another mode in the mixture of Gaussian model where fireballs are formed and shot. Whatever policy learned inside of this dream will achieve a perfect score of 2100 most of the time, but will obviously fail when unleashed into the harsh reality of the actual world, underperforming even a random policy.

Note again, however, that the simpler and more robust approach in \textit{Learning to Think} does not insist on using M for step by step planning. Instead, C can learn to use M's \textit{subroutines} (parts of M's weight matrix) for arbitrary computational purposes but can also learn to ignore M when M is useless and when ignoring M yields better performance. Nevertheless, at least in our present C--M variant, M's predictions are essential for teaching C, more like in some of the early C--M systems~\cite{s05_making_the_world_differentiable,s05a_cm,s05b_rl}, but combined with evolution or black box optimization.

By making the temperature $\tau$ an adjustable parameter of the M model, we can see the effect of training the C model on hallucinated virtual environments with different levels of uncertainty, and see how well they transfer over to the actual environment. We experimented with varying the temperature of the virtual environment and observing the resulting average score over 100 random rollouts of the actual environment after training the agent inside of the virtual environment with a given temperature:

\begin{table}[ht]
\label{doom_virtual_table}
\vskip -0.0in
\begin{center}
\begin{small}
\begin{sc}
\begin{tabular}{llll}
\toprule
Temperature $\tau$ & Virtual Score & Actual Score \\
\midrule
0.10 & 2086 $\pm$ 140 & 193 $\pm$ 58   \\
0.50 & 2060 $\pm$ 277 & 196 $\pm$ 50   \\
1.00 & 1145 $\pm$ 690 & 868 $\pm$ 511  \\
1.15 & 918 $\pm$ 546  & 1092 $\pm$ 556 \\
1.30 & 732 $\pm$ 269  & 753 $\pm$ 139  \\
\midrule
Random Policy & N/A & $210 \pm 108$ \\
Gym Leader & N/A & $820 \pm 58$ \\
\bottomrule
\end{tabular}
\end{sc}
\end{small}
\end{center}
\vskip -0.1in
\caption{\textit{Take Cover} scores at various temperature settings.}
\vskip -0.1in
\end{table}

We see that while increasing the temperature of the M model makes it more difficult for the C model to find adversarial policies, increasing it too much will make the virtual environment too difficult for the agent to learn anything, hence in practice it is a hyperparameter we can tune. The temperature also affects the types of strategies the agent discovers. For example, although the best score obtained is 1092 $\pm$ 556 with $\tau=1.15$, increasing $\tau$ a notch to 1.30 results in a lower score but at the same time a less risky strategy with a lower variance of returns. For comparison, the best score on the OpenAI Gym leaderboard~\cite{takecover} is 820 $\pm$ 58.
\vskip -0.1in
\section{Iterative Training Procedure}

In our experiments, the tasks are relatively simple, so a reasonable world model can be trained using a dataset collected from a random policy. But what if our environments become more sophisticated? In any difficult environment, only parts of the world are made available to the agent only after it learns how to strategically navigate through its world.

For more complicated tasks, an iterative training procedure is required. We need our agent to be able to explore its world, and constantly collect new observations so that its world model can be improved and refined over time. An iterative training procedure~\cite{learning_to_think} is as follows:

\begin{enumerate}
  \item Initialize M, C with random model parameters.
  \item Rollout to actual environment $N$ times. Save all actions $a_t$ and observations $x_t$ during rollouts to storage.
  \item Train M to model $P(x_{t+1}, r_{t+1}, a_{t+1}, d_{t+1} | x_t, a_t, h_t)$ and train C to optimize expected rewards inside of M.
  \item Go back to (2) if task has not been completed.
\end{enumerate}

We have shown that one iteration of this training loop was enough to solve simple tasks. For more difficult tasks, we need our controller in Step 2 to actively explore parts of the environment that is beneficial to improve its world model. An exciting research direction is to look at ways to incorporate artificial curiosity and intrinsic motivation~\cite{schmidhuber_creativity,s07_intrinsic,s08_curiousity,pathak2017,intrinsic_motivation} and information seeking~\cite{SchmidhuberStorck:94,Gottlieb2013} abilities in an agent to encourage novel exploration~\cite{Lehman2011}. In particular, we can augment the reward function based on improvement in compression quality~\cite{schmidhuber_creativity,s07_intrinsic,s08_curiousity,learning_to_think}.

In the present approach, since M is a MDN-RNN that models a probability distribution for the next frame, if it does a poor job, then it means the agent has encountered parts of the world that it is not familiar with. Therefore we can adapt and reuse M's training loss function to encourage curiosity. By flipping the sign of M's loss function in the actual environment, the agent will be encouraged to explore parts of the world that it is not familiar with. The new data it collects may improve the world model.

The iterative training procedure requires the M model to not only predict the next observation $x$ and $done$, but also predict the action and reward for the next time step. This may be required for more difficult tasks. For instance, if our agent needs to learn complex motor skills to walk around its environment, the world model will learn to imitate its own C model that has already learned to walk. After difficult motor skills, such as walking, is absorbed into a large world model with lots of capacity, the smaller C model can rely on the motor skills already absorbed by the world model and focus on learning more higher level skills to navigate itself using the motor skills it had already learned.

\begin{figure}[ht]
\vskip -0.0in
\begin{center}
\centerline{\includegraphics[width=1.0\columnwidth]{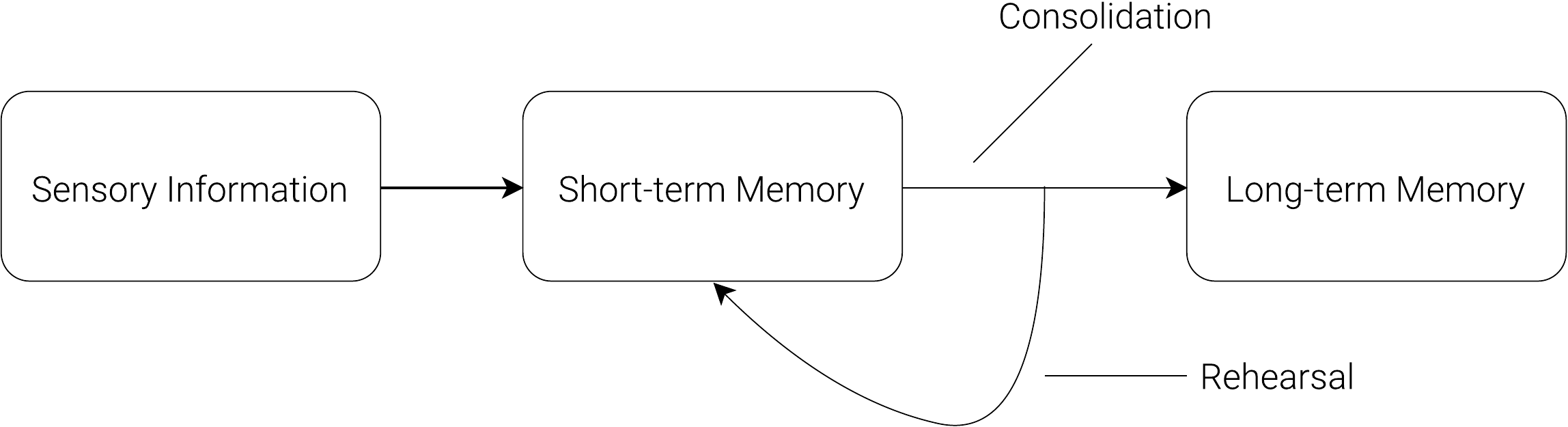}}
\caption{How information becomes memory.}
\vskip -0.1in
\end{center}
\vskip -0.2in
\end{figure}

An interesting connection to the neuroscience literature is the work on hippocampal replay that examines how the brain replays recent experiences when an animal rests or sleeps. Replaying recent experiences plays an important role in memory consolidation~\cite{Foster2017} -- where hippocampus-dependent memories become independent of the hippocampus over a period of time. As  \cite{Foster2017} puts it, replay is \textit{less like dreaming and more like thought}. We invite readers to read \textit{Replay Comes of Age}~\cite{Foster2017} for a detailed overview of replay from a neuroscience perspective with connections to theoretical reinforcement learning.

Iterative training could allow the C--M model to develop a natural hierarchical way to learn. Recent works about self-play in RL~\cite{asymmetric_self_play,competitive_self_play,continuous_adaptation_via_meta_learning} and PowerPlay~\cite{s10_powerplay,s11_powerplay} also explores methods that lead to a natural curriculum learning~\cite{s09_optimal_order}, and we feel this is one of the more exciting research areas of reinforcement learning.

\section{Related Work}

There is extensive literature on learning a dynamics model, and using this model to train a policy. Many concepts first explored in the 1980s for feed-forward neural networks (FNNs)~\cite{Werbos87specifications,Munro87,RobinsonFallside89,Werbos89identification,NguyenWidrow89} and in the 1990s for RNNs~\cite{s05_making_the_world_differentiable,s05a_cm,s05b_rl,s05c_boredom} laid some of the groundwork for \textit{Learning to Think}~\cite{learning_to_think}. The more recent PILCO~\cite{pilco,pilco_tutorial,McAllister2017} is a probabilistic model-based search policy method designed to solve difficult control problems. Using data collected from the environment, PILCO uses a Gaussian process (GP) model to learn the system dynamics, and then uses this model to sample many trajectories in order to train a controller to perform a desired task, such as swinging up a pendulum, or riding a unicycle.

\begin{figure}[ht]
\vskip -0.05in
\begin{center}
\centerline{\includegraphics[width=1.0\columnwidth]{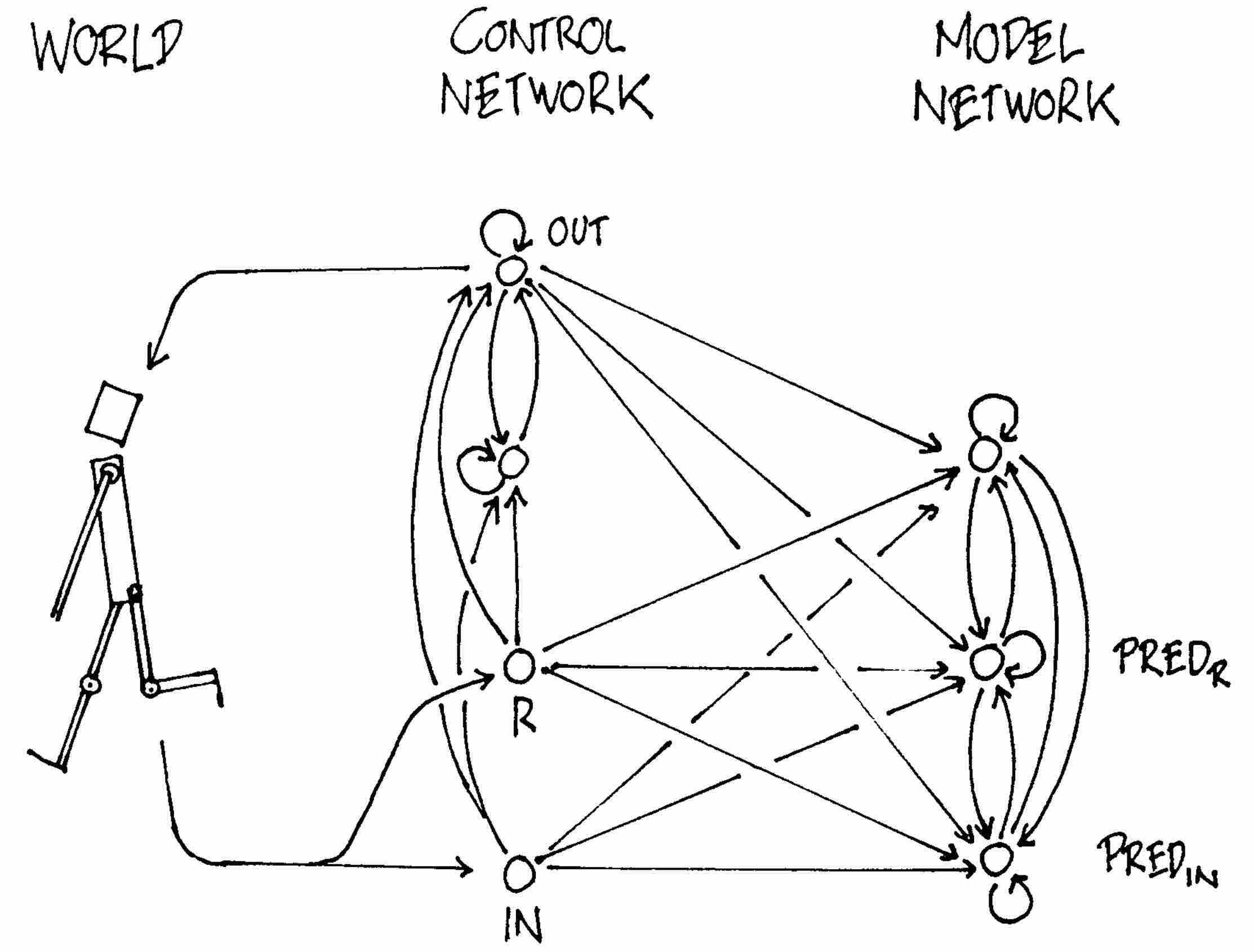}}
\vskip -0.10in
\caption{A controller with internal RNN model of the world \cite{s05_making_the_world_differentiable}.}
\end{center}
\vskip -0.25in
\end{figure}

While Gaussian processes work well with a small set of low dimension data, their computational complexity makes them difficult to scale up to model a large history of high dimensional observations. Other recent works~\cite{deep_pilco,Depeweg2017} use Bayesian neural networks instead of GPs to learn a dynamics model. These methods have demonstrated promising results on challenging control tasks~\cite{Hein2017}, where the states are known and well defined, and the observation is relatively low dimensional. Here we are interested in modelling dynamics observed from high dimensional visual data where our input is a sequence of raw pixel frames.

In robotic control applications, the ability to learn the dynamics of a system from observing only camera-based video inputs is a challenging but important problem. Early work on RL for active vision trained an FNN to take the current image frame of a video sequence to predict the next frame~\cite{s04_trajectories}, and use this predictive model to train a fovea-shifting control network trying to find targets in a visual scene. To get around the difficulty of training a dynamical model to learn directly from high-dimensional pixel images, researchers explored using neural networks to first learn a compressed representation of the video frames.  Recent work along these lines~\cite{learning_deep_dynamical_models_from_image_pixels,from_pixels_to_torques} was able to train controllers using the bottleneck hidden layer of an autoencoder as low-dimensional feature vectors to control a pendulum from pixel inputs. Learning a model of the dynamics from a compressed latent space enable RL algorithms to be much more data-efficient~\cite{deep_spacial_autoencoders,embed_to_control,finn_lecture}. We invite readers to watch Finn's lecture on Model-Based RL~\cite{finn_lecture} to learn more.

Video game environments are also popular in model-based RL research as a testbed for new ideas. \cite{game_engine_learning} used a feed-forward convolutional neural network (CNN) to learn a forward simulation model of a video game. Learning to predict how different actions affect future states in the environment is useful for game-play agents, since if our agent can predict what happens in the future given its current state and action, it can simply select the best action that suits its goal. This has been demonstrated not only in early work~\cite{NguyenWidrow89,s04_trajectories} (when compute was a million times more expensive than today) but also in recent studies~\cite{learn_to_act_by_predicting_future} on several competitive VizDoom environments.

The works mentioned above use FNNs to predict the next video frame. We may want to use models that can capture longer term time dependencies. RNNs are powerful models suitable for sequence modelling \cite{graves_rnn}. In a lecture called \textit{Hallucination with RNNs}~\cite{graves_lecture}, Graves demonstrated the ability of RNNs to learn a probabilistic model of Atari game environments. He trained RNNs to learn the structure of such a game and then showed that they can hallucinate similar game levels on its own.

Using RNNs to develop internal models to reason about the future has been explored as early as 1990 in a paper called \textit{Making the World Differentiable}~\cite{s05_making_the_world_differentiable}, and then further explored in \cite{s05a_cm,s05b_rl,s05c_boredom}. A more recent paper called \textit{Learning to Think}~\cite{learning_to_think} presented a unifying framework for building a RNN-based general problem solver that can learn a world model of its environment and also learn to reason about the future using this model. Subsequent works have used RNN-based models to generate many frames into the future~\cite{recurrent_env_sim,action_conditional_video_prediction,Denton2017}, and also as an internal model to reason about the future~\cite{Silver2016,imagination_agent,Watters2017}.

In this work, we used evolution strategies to train our controller, as it offers many benefits. For instance, we only need to provide the optimizer with the final cumulative reward, rather than the entire history. ES is also easy to parallelize -- we can launch many instances of \texttt{rollout} with different solutions to many workers and quickly compute a set of cumulative rewards in parallel. Recent works \cite{pathnet,openai,stablees,stanley2017} have confirmed that ES is a viable alternative to traditional Deep RL methods on many strong baselines.

Before the popularity of Deep RL methods~\cite{dqn}, evolution-based algorithms have been shown to be effective at solving RL tasks~\cite{neat,gom5_ne_accelerated,gom2_coevolve,hyperneat,pepg,evolving_neural_networks}. Evolution-based algorithms have even been able to solve difficult RL tasks from high dimensional pixel inputs~\cite{kou1_torcs,hausknecht,parker2012}. More recent works~\cite{vae_evolution} combine VAE and ES, which is similar to our approach.

\section{Discussion}

\begin{figure}[ht]
\vskip -0.05in
\begin{center}
\centerline{\includegraphics[width=1.0\columnwidth]{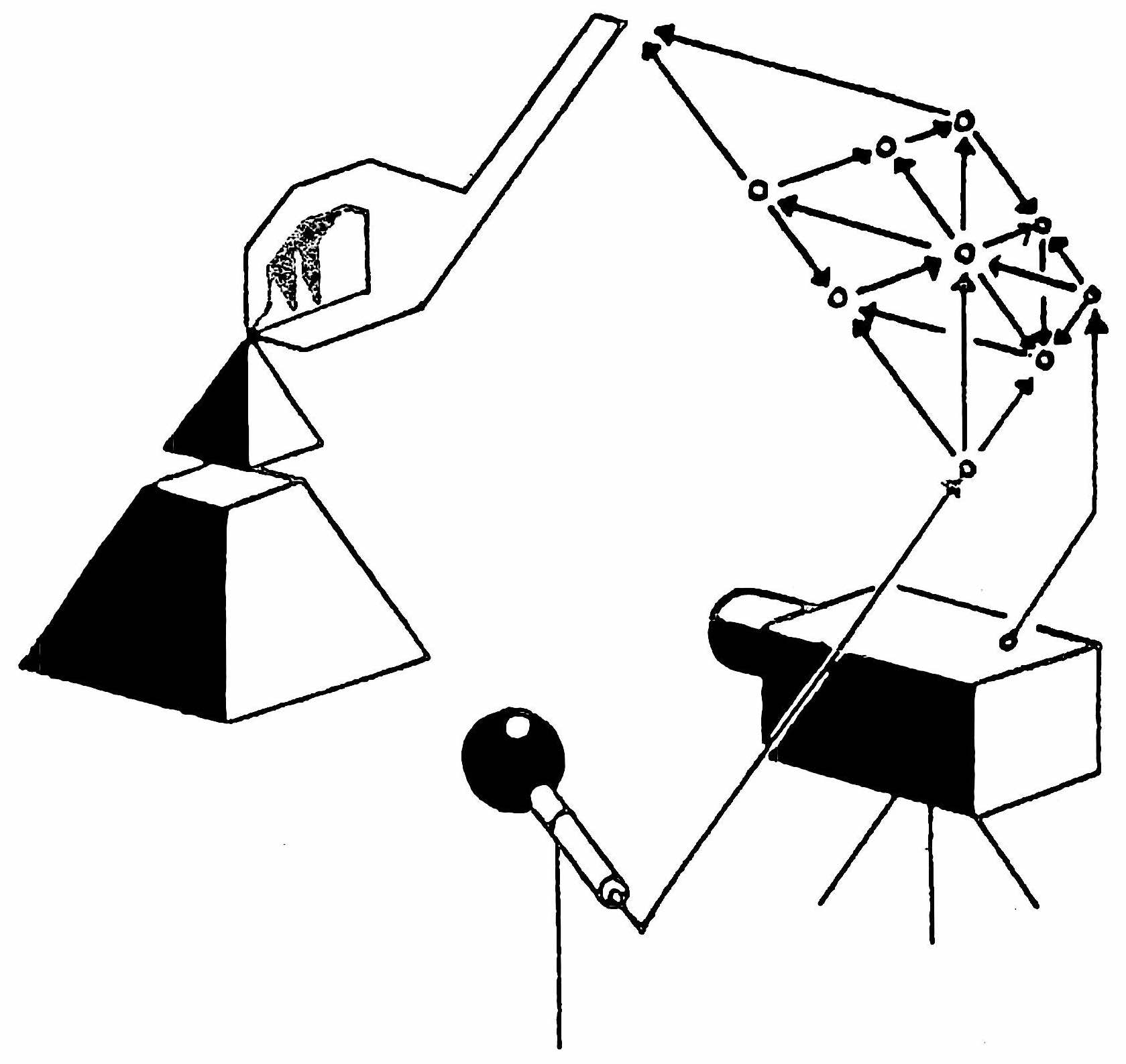}}
\caption{Ancient drawing (1990) of a RNN-based controller interacting with an environment \cite{s05_making_the_world_differentiable}.}
\vskip -0.15in
\end{center}
\vskip -0.1in
\end{figure}

We have demonstrated the possibility of training an agent to perform tasks entirely inside of its simulated latent space dream world. This approach offers many practical benefits. For instance, running computationally intensive game engines require using heavy compute resources for rendering the game states into image frames, or calculating physics not immediately relevant to the game. We may not want to waste cycles training an agent in the actual environment, but instead train the agent as many times as we want inside its simulated environment. Training agents in the real world is even more expensive, so world models that are trained incrementally to simulate reality may prove to be useful for transferring policies back to the real world. Our approach may complement \textit{sim2real} approaches outlined in \cite{Bousmalis2017,Higgins2017}.

Furthermore, we can take advantage of deep learning frameworks to accelerate our world model simulations using GPUs in a distributed environment. The benefit of implementing the world model as a fully differentiable recurrent computation graph also means that we may be able to train our agents in the dream directly using the backpropagation algorithm to fine-tune its policy to maximize an objective function \cite{s05_making_the_world_differentiable,s05a_cm,s05b_rl}.

The choice of using a VAE for the V model and training it as a standalone model also has its limitations, since it may encode parts of the observations that are not relevant to a task. After all, unsupervised learning cannot, by definition, know what will be useful for the task at hand. For instance, it reproduced unimportant detailed brick tile patterns on the side walls in the Doom environment, but failed to reproduce task-relevant tiles on the road in the Car Racing environment. By training together with a M model that predicts rewards, the VAE may learn to focus on task-relevant areas of the image, but the tradeoff here is that we may not be able to reuse the VAE effectively for new tasks without retraining.

Learning task-relevant features has connections to neuroscience as well. Primary sensory neurons are released from inhibition when rewards are received, which suggests that they generally learn task-relevant features, rather than just any features, at least in adulthood~\cite{Pi2013}.


Another concern is the limited capacity of our world model. While modern storage devices can store large amounts of historical data generated using the iterative training procedure, our LSTM~\cite{lstm,s12_lstm_forget}-based world model may not be able to store all of the recorded information inside its weight connections. While the human brain can hold decades and even centuries of memories to some resolution~\cite{brain_capacity}, our neural networks trained with backpropagation have more limited capacity and suffer from issues such as catastrophic forgetting~\cite{Ratcliff1990,French1994,Kirkpatrick2016}. Future work may explore replacing the small MDN-RNN network with higher capacity models ~\cite{outrageously_large_neural_nets,hypernetworks,suarez2017,wavenet,attention}, or incorporating an external memory module~\cite{Gemici2017}, if we want our agent to learn to explore more complicated worlds.

Like early RNN-based C--M systems ~\cite{s05_making_the_world_differentiable,s05a_cm,s05b_rl,s05c_boredom}, ours simulates possible futures time step by time step, without profiting from human-like hierarchical planning or abstract reasoning, which often ignores irrelevant spatial-temporal details. However, the more general \textit{Learning To Think}~\cite{learning_to_think} approach is not limited to this rather naive approach. Instead it allows a recurrent C to learn to address \textit{subroutines} of the recurrent M, and reuse them for problem solving in arbitrary computable ways, e.g., through hierarchical planning or other kinds of exploiting parts of M's program-like weight matrix. A recent \textit{One Big Net}~\cite{onebignet2018} extension of the C--M approach
collapses C and M into a single network, and uses PowerPlay-like~\cite{s10_powerplay,s11_powerplay} behavioural replay (where the behaviour of a teacher net is compressed into a student net~\cite{chunker91and92}) to avoid forgetting old prediction and control skills when learning new ones. Experiments with those more general approaches are left for future work.

\section*{Acknowledgements}

We would like to thank Blake Richards, Kai Arulkumaran, Ankur Handa, Kory Mathewson, Kyle McDonald, Denny Britz, Elwin Ha and Natasha Jaques for their thoughtful feedback on this article, and for offering their valuable perspectives and insights from their areas of expertise.

The interactive online version of this article was built using \url{distill.pub}'s web technology. We would like to thank Chris Olah and the rest of the Distill editorial team for their valuable feedback and generous editorial support, in addition to supporting the use of their Distill technology.

The interative demos on \url{worldmodels.github.io} were all built using \texttt{p5.js}. Deploying all of these machine learning models in a web browser was made possible with \texttt{deeplearn.js}, a hardware-accelerated machine learning framework for the browser, developed by the \textit{People+AI Research Initiative} (PAIR) team at Google. A special thanks goes to Nikhil Thorat and Daniel Smilkov for their help during the development process.

We would to extend our thanks to Alex Graves, Douglas Eck, Mike Schuster, Rajat Monga, Vincent Vanhoucke, Jeff Dean and the Google Brain team for helpful feedback and for encouraging us to explore this area of research. Experiments were performed on Ubuntu virtual machines provided by Google Cloud Platform. Any errors here are our own and do not reflect opinions of our proofreaders and colleagues.

%









\appendix
\section{Appendix}

In this section we will describe in more details the models and training methods used in this work.

\subsection{Variational Autoencoder}

\begin{figure}[ht]
\vskip -0.10in
\begin{center}
\centerline{\includegraphics[width=0.5\columnwidth]{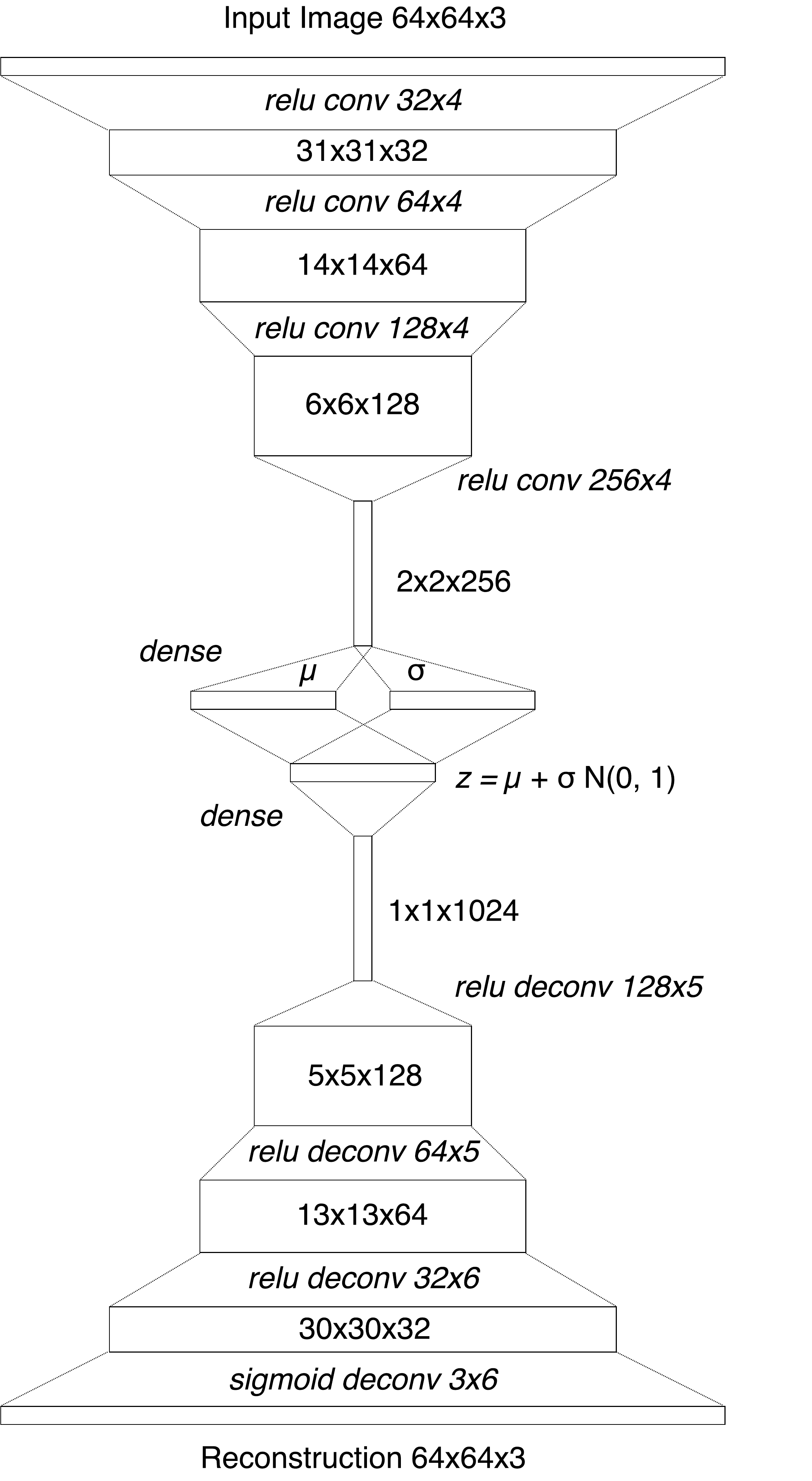}}
\vskip -0.15in
\caption{Description of tensor shapes at each layer of ConvVAE.}
\end{center}
\vskip -0.30in
\end{figure}

We trained a Convolutional Variational Autoencoder (ConvVAE) model as the V Model of our agent. Unlike vanilla autoencoders, enforcing a Gaussian prior over the latent vector $z$ also limits the amount of information capacity for compressing each frame, but this Gaussian prior also makes the world model more robust to unrealistic $z$ vectors generated by the M Model.

As the environment may give us observations as high dimensional pixel images, we first resize each image to 64x64 pixels before and use this resized image as the V Model's observation. Each pixel is stored as three floating point values between 0 and 1 to represent each of the RGB channels. The ConvVAE takes in this 64x64x3 input tensor and passes this data through 4 convolutional layers to encode it into low dimension vectors $\mu$ and $\sigma$, each of size $N_z$. The latent vector $z$ is sampled from the Gaussian prior $N(\mu, \sigma I)$. In the Car Racing task, $N_z$ is 32 while for the Doom task $N_z$ is 64. The latent vector $z$ is passed through 4 of deconvolution layers used to decode and reconstruct the image.

Each convolution and deconvolution layer uses a stride of 2. The layers are indicated in the diagram in \textit{Italics} as \textit{Activation-type Output Channels x Filter Size}. All convolutional and deconvolutional layers use relu activations except for the output layer as we need the output to be between 0 and 1. We trained the model for 1 epoch over the data collected from a random policy, using $L^2$ distance between the input image and the reconstruction to quantify the reconstruction loss we optimize for, in addition to KL loss.

\subsection{Recurrent Neural Network}

For the M Model, we use an LSTM~\cite{lstm} recurrent neural network combined with a Mixture Density Network~\cite{bishop} as the output layer. We use this network to model the probability distribution of the next $z$ in the next time step as a Mixture of Gaussian distribution. This approach is very similar to \cite{graves_rnn} in the Unconditional Handwriting Generation section and also the decoder-only section of SketchRNN~\cite{sketchrnn}. The only difference in the approach used is that we did not model the correlation parameter between each element of $z$, and instead had the MDN-RNN output a diagonal covariance matrix of a factored Gaussian distribution.

\begin{figure}[ht]
\vskip -0.05in
\begin{center}
\centerline{\includegraphics[width=1.0\columnwidth]{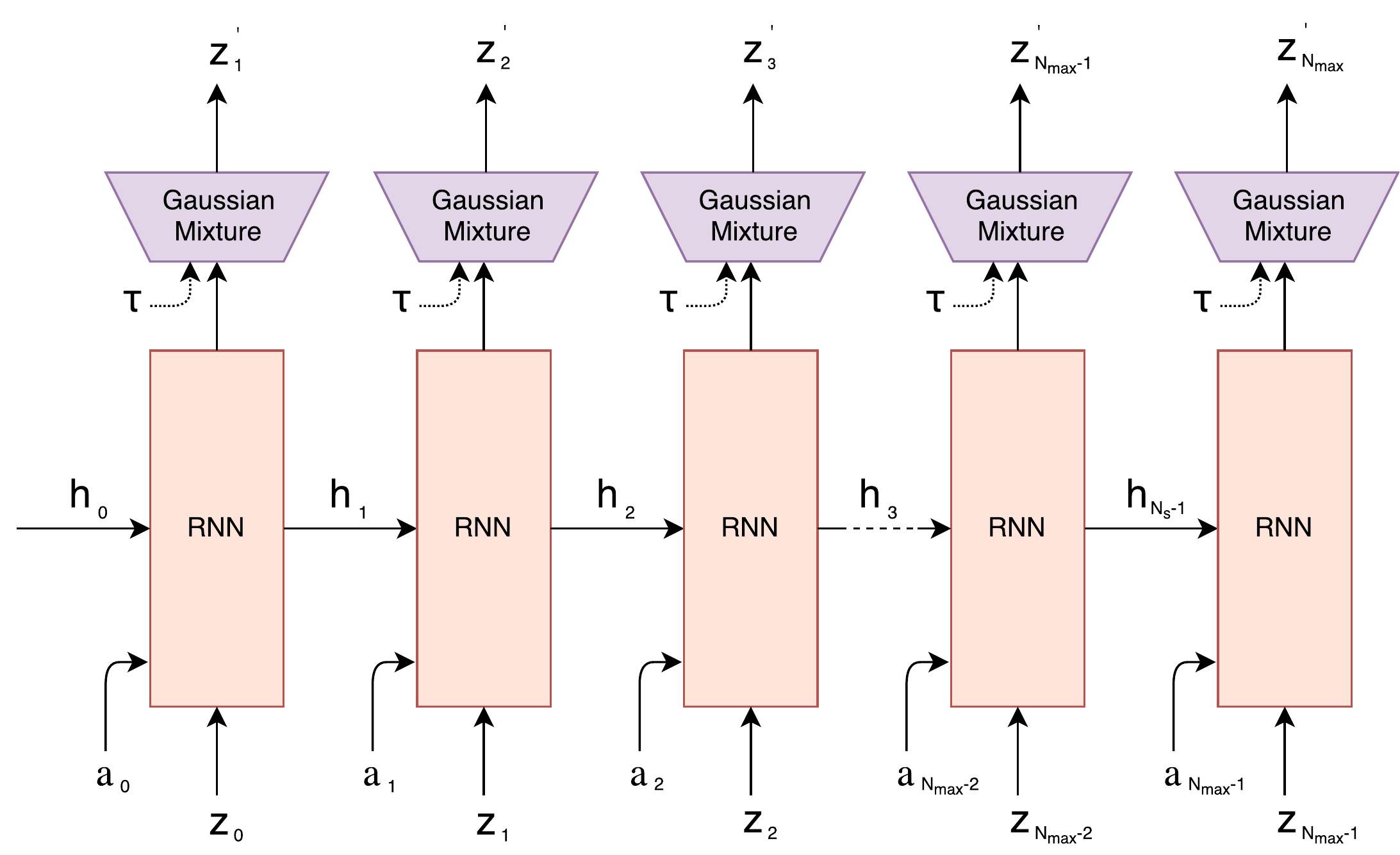}}
\vskip -0.00in
\caption{MDN-RNN decoder similar to \cite{graves_rnn,sketchrnn}}
\end{center}
\vskip -0.35in
\end{figure}

Unlike the handwriting and sketch generation works, rather than using the MDN-RNN to model the pdf of the next pen stroke, we model instead the pdf of the next latent vector $z$. We would sample from this pdf at each time step to generate the hallucinated environments. In the Doom task, we also also use the MDN-RNN to predict the probability of whether the agent has died in this frame. If that probability is above 50\%, then we set \textit{done} to be \textit{true} in the virtual environment. Given that death is a low probability event at each time step, we find the cutoff approach to more stable compared to sampling from the Bernoulli distribution.

The MDN-RNNs were trained for 20 epochs on the data collected from a random policy agent. In the Car Racing task, the LSTM used 256 hidden units, while the Doom task used 512 hidden units. In both tasks, we used 5 Gaussian mixtures and did not model the correlation $\rho$ parameter, hence $z$ is sampled from a factored mixture of Gaussian distribution.

When training the MDN-RNN using teacher forcing from the recorded data, we store a pre-computed set of $\mu$ and $\sigma$ for each of the frames, and sample an input $z \sim N(\mu, \sigma)$ each time we construct a training batch, to prevent overfitting our MDN-RNN to a specific sampled $z$.

\subsection{Controller}

For both environments, we applied $\tanh$ nonlinearities to clip and bound the action space to the appropriate ranges. For instance, in the Car Racing task, the steering wheel has a range from -1 to 1, the acceleration pedal from 0 to 1, and the brakes from 0 to 1. In the Doom environment, we converted the discrete actions into a continuous action space between -1 to 1, and divided this range into thirds to indicate whether the agent is moving left, staying where it is, or moving to the right. We would give the C Model a feature vector as its input, consisting of $z$ and the hidden state of the MDN-RNN. In the Car Racing task, this hidden state is the output vector $h$ of the LSTM, while for the Doom task it is both the cell vector $c$ and the output vector $h$ of the LSTM.

\subsection{Evolution Strategies}

We used \textit{Covariance-Matrix Adaptation Evolution Strategy} (CMA-ES)~\cite{cmaes} to evolve the weights for our C Model. Following the approach described in \textit{Evolving Stable Strategies} \cite{stablees}, we used a population size of 64, and had each agent perform the task 16 times with different initial random seeds. The fitness value for the agent is the \textit{average cumulative reward} of the 16 random rollouts. The diagram below charts the best performer, worst performer, and mean fitness of the population of 64 agents at each generation:

\begin{figure}[ht]
\vskip -0.00in
\begin{center}
\centerline{\includegraphics[width=1.0\columnwidth]{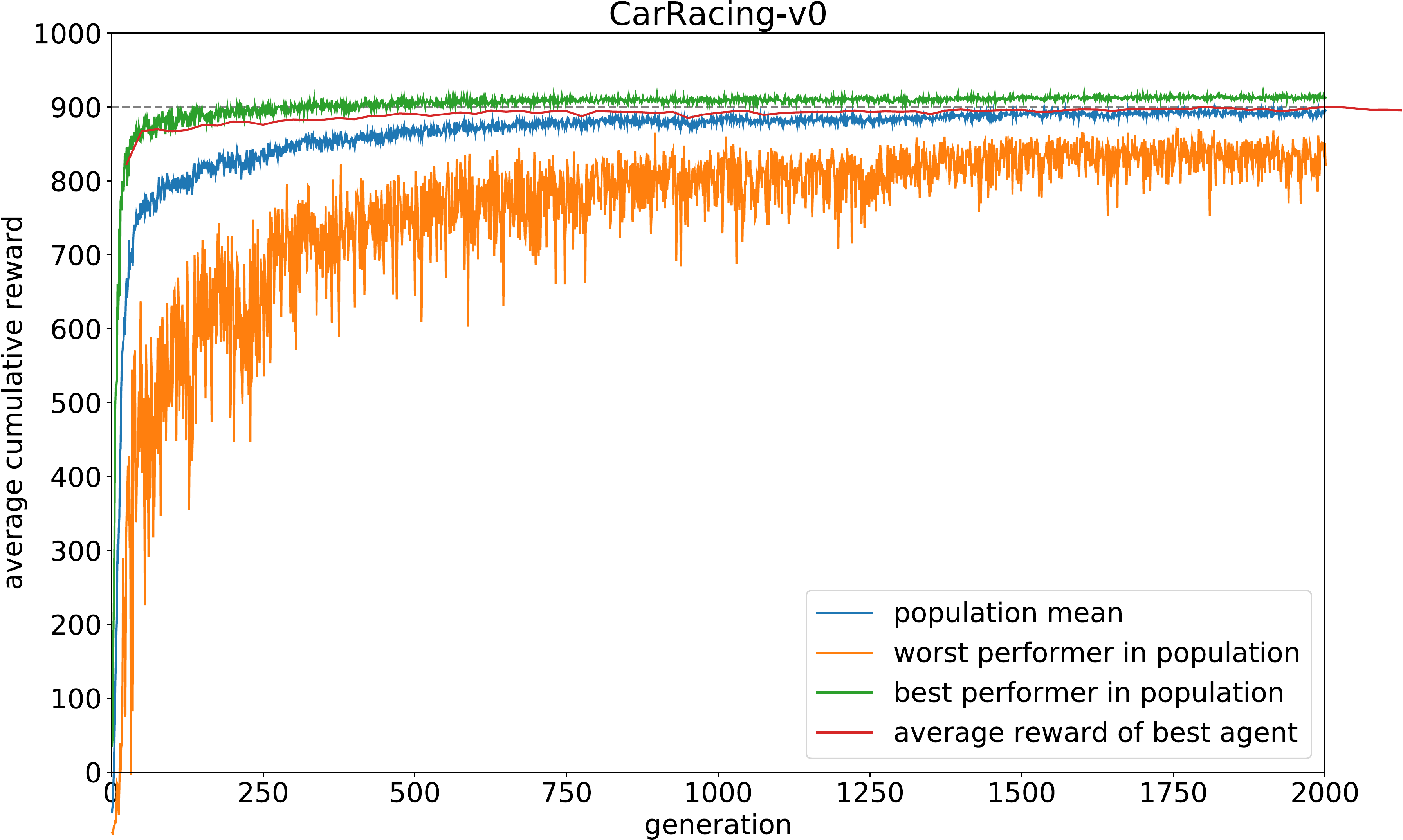}}
\vskip -0.15in
\caption{Training of \texttt{CarRacing-v0}}
\end{center}
\vskip -0.25in
\end{figure}

Since the requirement of this environment is to have an agent achieve an average score above 900 over 100 random rollouts, we took the best performing agent at the end of every 25 generations, and tested that agent over 1024 random rollout scenarios to record this average on the red line. After 1800 generations, an agent was able to achieve an average score of 900.46 over 1024 random rollouts. We used 1024 random rollouts rather than 100 because each process of the 64 core machine had been configured to run 16 times already, effectively using a full generation of compute after every 25 generations to evaluate the best agent 1024 times. Below, we plot the results of same agent evaluated over 100 rollouts:

\begin{figure}[ht]
\vskip -0.05in
\begin{center}
\centerline{\includegraphics[width=1.0\columnwidth]{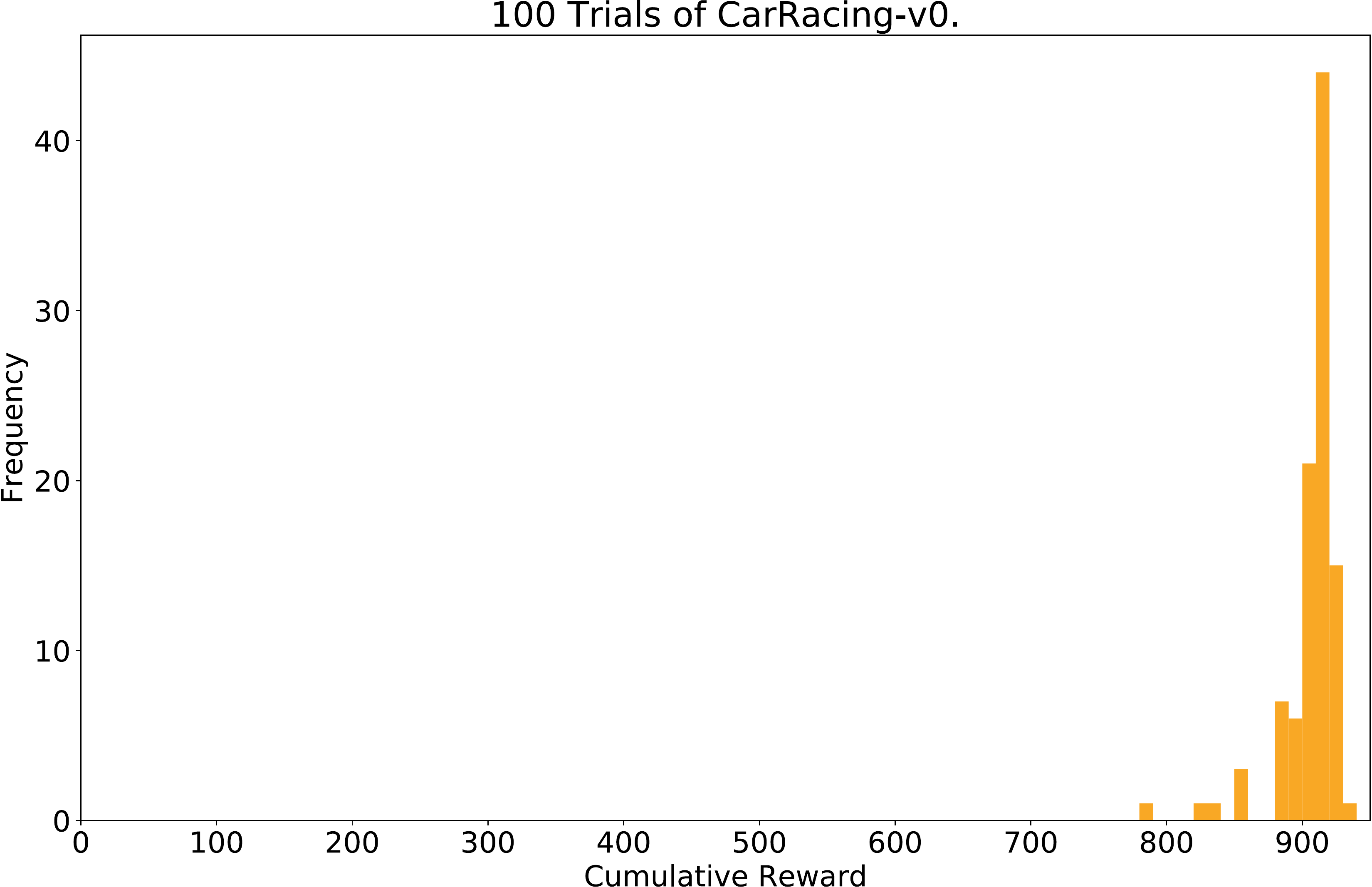}}
\vskip -0.10in
\caption{Histogram of cumulative rewards. Score is 906 $\pm$ 21.}
\end{center}
\vskip -0.4in
\end{figure}

We also experimented with an agent that has access to only the $z$ vector from the VAE, and not letting it see the RNN's hidden states. We tried 2 variations, where in the first variation, the C Model mapped $z$ directly to the action space $a$. In second variation, we attempted to add a hidden layer with 40 $tanh$ activations between $z$ and $a$, increasing the number of model parameters of the C Model to 1443, making it more comparable with the original setup.

\begin{figure}[ht]
\vskip -0.05in
\begin{center}
\centerline{\includegraphics[width=1.0\columnwidth]{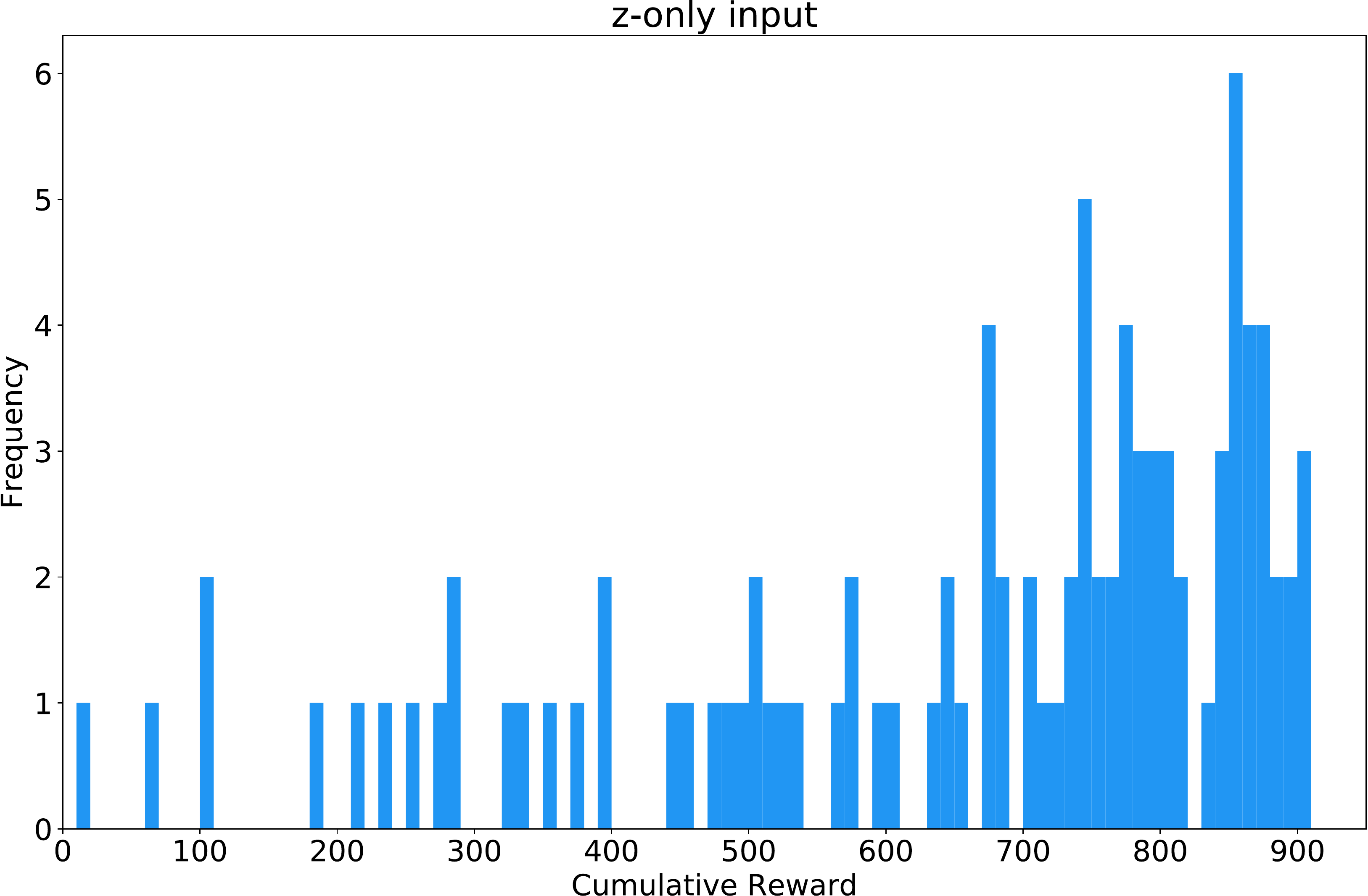}}
\vskip -0.10in
\caption{When agent sees only $z_t$, score is 632 $\pm$ 251.}
\end{center}
\vskip -0.4in
\end{figure}

\begin{figure}[ht]
\vskip -0.05in
\begin{center}
\centerline{\includegraphics[width=1.0\columnwidth]{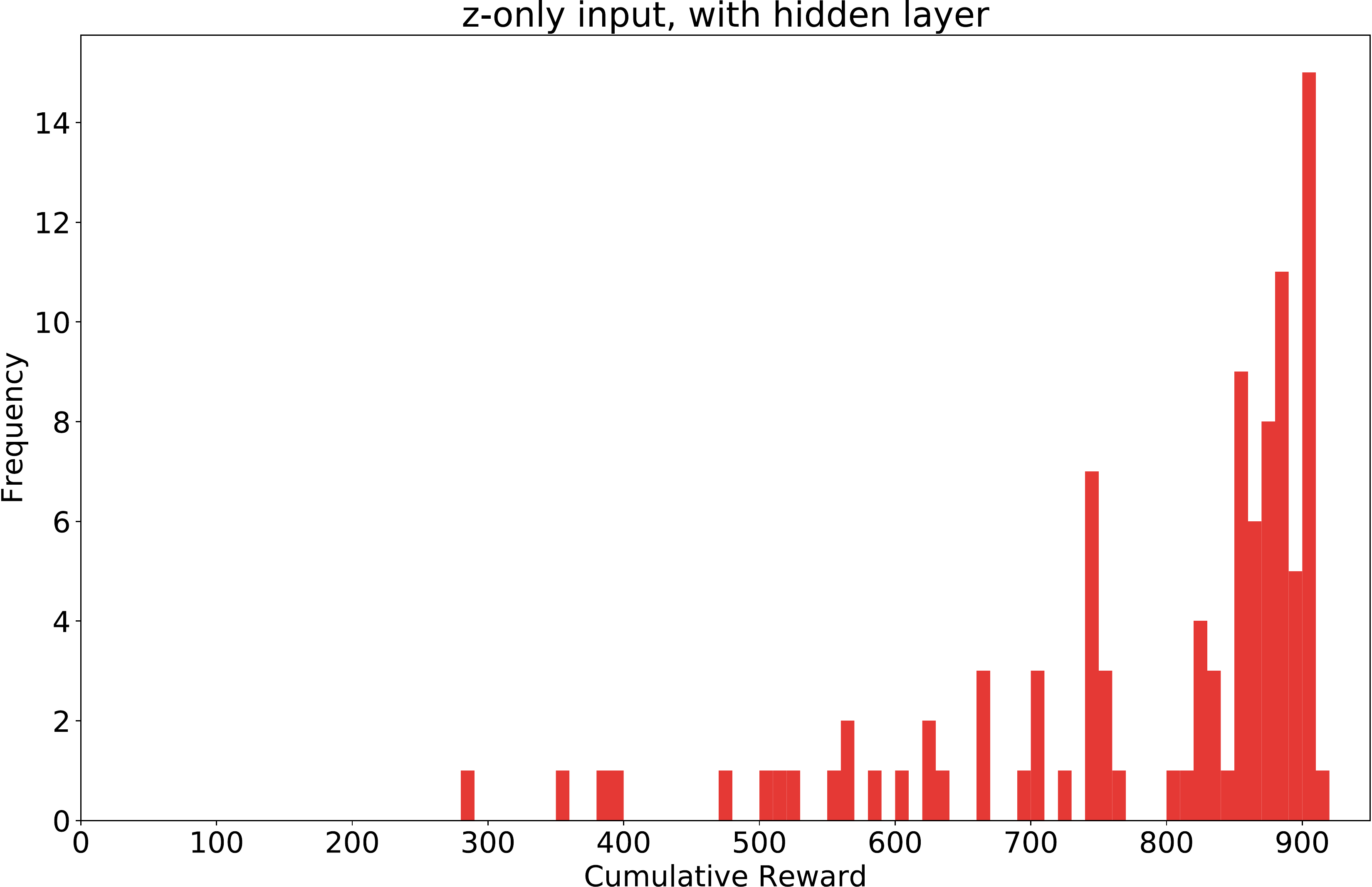}}
\vskip -0.10in
\caption{When agent sees only $z_t$, with a hidden layer, score is 788 $\pm$ 141.}
\end{center}
\vskip -0.4in
\end{figure}

\subsection{DoomRNN}

We conducted a similar experiment on the hallucinated Doom environment we called \textit{DoomRNN}. Please note that we have not actually attempted to train our agent on the actual VizDoom environment, and had only used VizDoom for the purpose of collecting training data using a random policy. \textit{DoomRNN} is more computationally efficient compared to VizDoom as it only operates in latent space without the need to render a screenshot at each time step, and does not require running the actual Doom game engine.

\begin{figure}[ht]
\vskip -0.05in
\begin{center}
\centerline{\includegraphics[width=1.0\columnwidth]{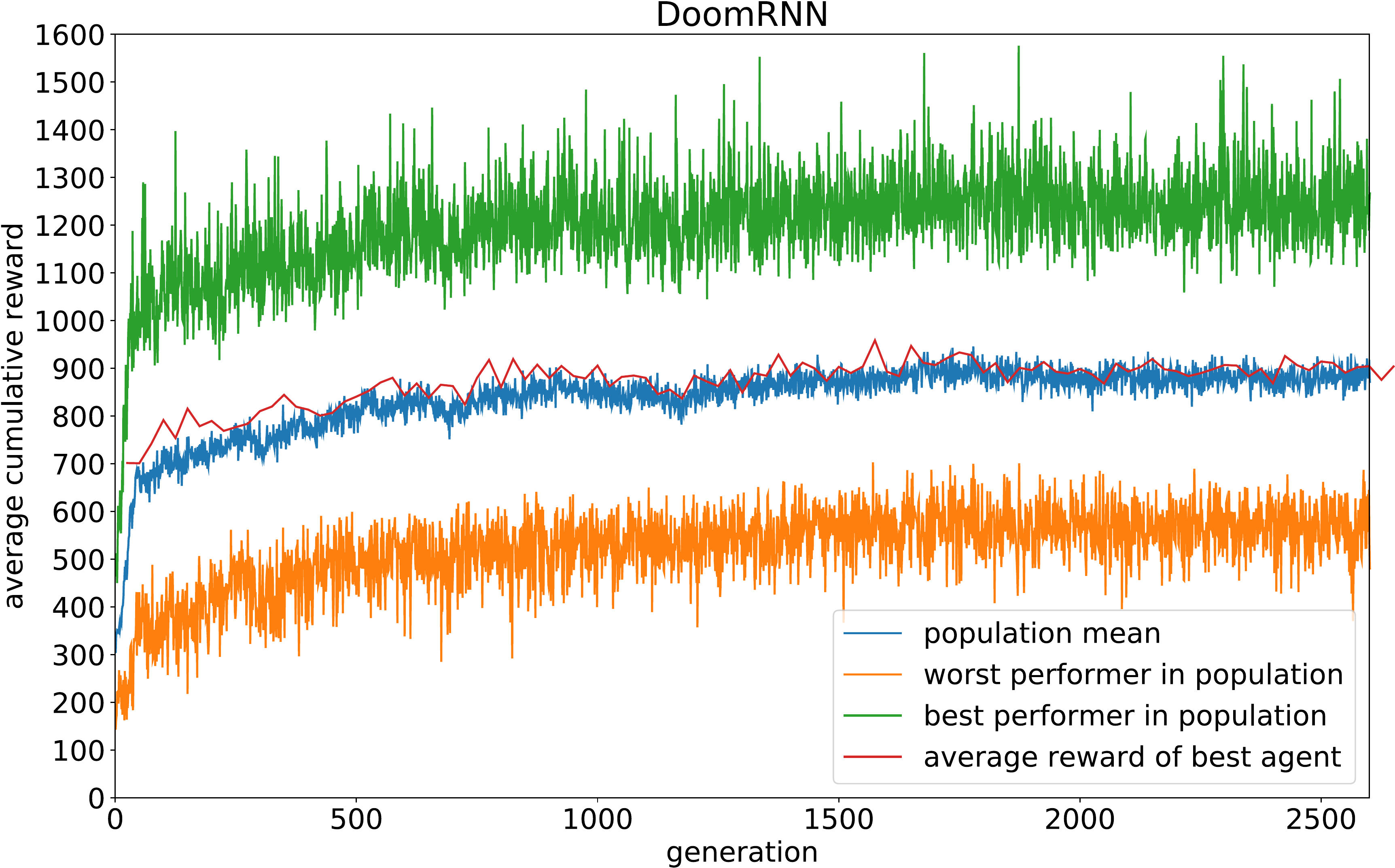}}
\vskip -0.05in
\caption{Training of DoomRNN.}
\end{center}
\vskip -0.2in
\end{figure}

In the virtual DoomRNN environment we constructed, we increased the temperature slightly and used $\tau=1.15$ to make the agent learn in a more challenging environment. The best agent managed to obtain an average score of 959 over 1024 random rollouts (the highest score of the red line in the diagram). This same agent achieved an average score of 1092 $\pm$ 556 over 100 random rollouts when deployed to the actual \textit{DoomTakeCover-v0}~\cite{takecover} environment.

\begin{figure}[ht]
\vskip -0.00in
\begin{center}
\centerline{\includegraphics[width=0.9\columnwidth]{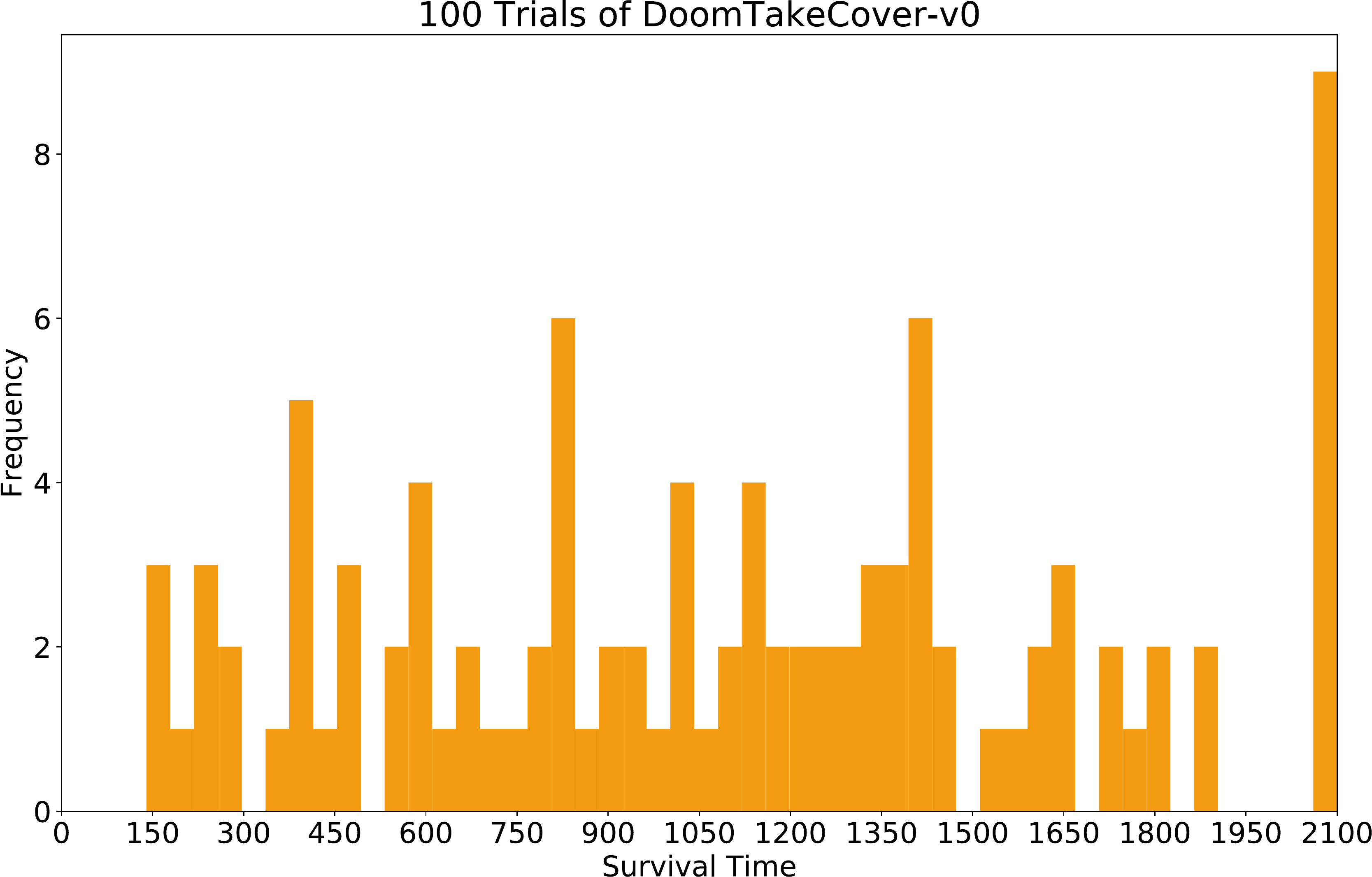}}
\vskip -0.05in
\caption{Histogram of time steps survived in the actual VizDoom environment over 100 consecutive trials. Score is 1092 $\pm$ 556.}
\end{center}
\vskip -0.2in
\end{figure}

\bibliography{main}
\bibliographystyle{icml2018}

\end{document}